%% file: root.tex
\documentclass[letterpaper,journal]{IEEEtran}
\usepackage{amsmath,amsfonts}
\usepackage{array}
\usepackage[caption=false,font=footnotesize]{subfig}

\usepackage{textcomp}
\usepackage{stfloats}
\usepackage{url}
\usepackage{verbatim}
\usepackage{graphicx}
\usepackage{cite}
\usepackage[OT1]{fontenc}

\graphicspath{{./images/front_figure/}{./figures/}{./}}

\usepackage{amssymb}
\usepackage{makecell}
\usepackage{multirow}
\usepackage{hyperref}
\hypersetup{
  colorlinks    = false, %
  urlcolor      = blue, %
  linkcolor     = blue, %
  citecolor     = red   %
}

\usepackage{siunitx}

\usepackage{pifont}
\newcommand{\cmark}{\ding{51}}%
\newcommand{\xmark}{\ding{55}}%

\usepackage{tikz}
\usetikzlibrary{arrows.meta, positioning, calc, fit, backgrounds}
\usepackage{pgfplots}
\usepackage{pgfgantt}
\usepackage{pgf-pie} %
\usepackage{xcolor}

\usepackage{algorithm}
\makeatletter
\providecommand\expanded[1]{\edef\@tempa{#1}\@tempa}
\makeatother
\usepackage{algpseudocodex}

\usepackage{transparent}

\usepackage{multirow}
\usepackage{rotating}
\usepackage{bbding}
\usepackage{diagbox}

\input{edited.tex}

\input{gloss}

\DeclareMathOperator*{\argmin}{argmin}

\def\videolink{\href{http://tiny.cc/sem-explor-semi-static}{http://tiny.cc/sem-explor-semi-static}}
\def\webpagelink{\href{https://utiasdsl.github.io/semi-static-semantic-exploration/}{https://utiasdsl.github.io/semi-static-semantic-exploration/}}

\makeatletter
\define@key{Gin}{Trim}
           {\let\Gin@viewport@code\Gin@trim\expandafter\Gread@parse@vp#1 \\}
\makeatother

\begin{document}

\title{Where Did I Leave My Glasses? Open-Vocabulary Semantic Exploration in Real-World Semi-Static Environments}

\author{\coloredited{}Benjamin Bogenberger, Oliver Harrison, Orrin Dahanaggamaarachchi, Lukas Brunke, Jingxing Qian,\\Siqi Zhou, Angela P. Schoellig%
\thanks{\coloredited{}This work was supported by the German Federal Ministry of Research, Technology and Space (BMFTR) under the Robotics Institute Germany (RIG) with BMBF grant 16ME0997K and the EU's Horizon Europe project under the Marie Skłodowska-Curie Actions grant agreement No. 101155035.}%
\thanks{\coloredited{}The authors are with the Learning Systems and Robotics Lab and the Munich Institute of Robotics and Machine Intelligence, Technical University of Munich, 80333 Munich, Germany. Lukas Brunke, Jingxing Qian, and Angela P. Schoellig are also affiliated with the University of Toronto Institute for Aerospace Studies, the University of Toronto Robotics Institute, and the Vector Institute for Artificial Intelligence, Toronto, Canada. Siqi Zhou is currently with Simon Fraser University, Vancouver, Canada. Email: {\small\tt firstname.lastname@tum.de}}%
}

\maketitle
\thispagestyle{empty}
\pagestyle{empty}

\input{sections/abstract}


\input{sections/introduction}

\input{sections/related_work}

\input{sections/methodology}

\input{sections/evaluation}

\input{sections/conclusion}

\bibliographystyle{IEEEtran}

\end{document}

%% file: edited.tex
\newif\ifinedited
\ineditedfalse

\let\origtable\table
\let\endorigtable\endtable

\let\origfigure\figure
\let\endorigfigure\endfigure

\def\editc{black}

\renewenvironment{table}
  {\ifinedited \origtable\color{\editc} \else \origtable \fi}
  {\endorigtable}

\renewenvironment{figure}
  {\ifinedited \origfigure\color{\editc} \else \origfigure \fi}
  {\endorigfigure}

\def\coloredited{}

%% file: gloss.tex
\usepackage[acronyms, style=alttree, nonumberlist, nopostdot, shortcuts]{glossaries}

\glsdisablehyper

\makenoidxglossaries

\glssetwidest{THISWI}

\newacronym[description=Model predictive control]{mpc}{MPC}{model predictive control}
\newacronym[description=Large language model]{llm}{LLM}{large language model}
\newacronym[description=Reinforcement learning]{rl}{RL}{reinforcement learning}
\newacronym[description=Simoultanous mapping and localization]{slam}{SLAM}{simoultanous mapping and localization}
\newacronym[description=Neureal radiance field]{nerf}{NERF}{neural radiance field}
\newacronym[description=Gaussian splatting]{gs}{GS}{Gaussian splatting}
\newacronym[description=Visual inertia odometry]{vio}{VIO}{visual inertia odometry}
\newacronym[description=Root mean square error]{rmse}{RMSE}{root mean square error}
\newacronym[description=iterative closest point]{icp}{ICP}{iterative closest point}
\newacronym[description=degree of freedom]{dof}{DOF}{degree of freedom}
\newacronym[description=success weighted by path length]{spl}{SPL}{success weighted by path length}

\renewcommand{\vec}[1]{\mathbf{\MakeLowercase{#1}}}
\newcommand{\Mat}[1]{\mathbf{\MakeUppercase{#1}}}

\newcommand{\Symbols}{List of Symbols}
\newglossary{symbols}{sym}{sbl}{\Symbols}
\newglossaryentry{frames}{type=symbols,
	sort={},
	name={\ensuremath{\mathcal{F}}},
	description={set of frames}
}
\newglossaryentry{frame}{type=symbols,
	sort={},
	name={\ensuremath{\Mat{F}}},
	description={frame}
}

\newglossaryentry{tfs_cam}{type=symbols,
	sort={},
	name={\ensuremath{\mathcal{T}^\mathrm{CW}}},
	description={set of camera-to-world transformations}
}
\newglossaryentry{tf_cam}{type=symbols,
	sort={},
	name={\ensuremath{\Mat{T}^\mathrm{CW}}},
	description={camera-to-world transformation}
}

\newglossaryentry{obj_lib}{type=symbols,
	sort={},
	name={\ensuremath{\mathcal{O}}},
	description={Object library}
}
\newglossaryentry{obj}{type=symbols,
	sort={},
	name={\ensuremath{\Mat{O}}},
	description={Object}
}

\newglossaryentry{obj_tf}{type=symbols,
	sort={},
	name={\ensuremath{\Mat{T}^\mathrm{OW}}},
	description={Object-to-camera transformation},
	parent=obj
}
\newglossaryentry{obj_pc}{type=symbols,
	sort={},
	name={\ensuremath{\Mat{P}}},
	description={Object pointcloud},
	parent=obj
}
\newglossaryentry{obj_b}{type=symbols,
	sort={},
	name={\ensuremath{\Mat{B}}},
	description={Object bounding box},
	parent=obj
}
\newglossaryentry{obj_f}{type=symbols,
	sort={},
	name={\ensuremath{\vec{f}}},
	description={Object visual feature},
	parent=obj
}
\newglossaryentry{obj_c}{type=symbols,
	sort={},
	name={\ensuremath{\vec{c}}},
	description={Object open-vocabulary class},
	parent=obj
}
\newglossaryentry{obj_s}{type=symbols,
	sort={},
	name={\ensuremath{\vec{s}}},
	description={Object stationarity class},
	parent=obj
}
\newglossaryentry{obj_t_first}{type=symbols,
	sort={},
	name={\ensuremath{t^+}},
	description={Object first observed time},
	parent=obj
}
\newglossaryentry{obj_t_disappear}{type=symbols,
	sort={},
	name={\ensuremath{t^-}},
	description={Object disappeared time},
	parent=obj
}

\newglossaryentry{det_tf}{type=symbols,
	sort={},
	name={\ensuremath{\hat{\Mat{T}}^\mathrm{OW}}},
	description={Observation-to-camera transformation},
	parent=obj
}
\newglossaryentry{det_pc}{type=symbols,
	sort={},
	name={\ensuremath{\hat{\Mat{P}}}},
	description={Observation pointcloud},
	parent=obj
}
\newglossaryentry{det_b}{type=symbols,
	sort={},
	name={\ensuremath{\hat{\Mat{B}}}},
	description={Observation bounding box},
	parent=obj
}
\newglossaryentry{det_f}{type=symbols,
	sort={},
	name={\ensuremath{\hat{\vec{f}}}},
	description={Observation visual feature},
	parent=obj
}
\newglossaryentry{det_c}{type=symbols,
	sort={},
	name={\ensuremath{\hat{\vec{c}}}},
	description={Observation open-vocabulary class},
	parent=obj
}

\newglossaryentry{detections}{type=symbols,
	sort={},
	name={\ensuremath{\mathcal{Y}}},
	description={set of detections}
}
\newglossaryentry{detection}{type=symbols,
	sort={},
	name={\ensuremath{\Mat{Y}}},
	description={detection}
}

\newglossaryentry{geom_change}{type=symbols,
	sort={},
	name={\ensuremath{l}},
	description={geoemetric change},
}
\newglossaryentry{pocd_score}{type=symbols,
	sort={},
	name={\ensuremath{v}},
	description={stationarityy score},
}
\newglossaryentry{change}{type=symbols,
	sort={},
	name={\ensuremath{\Delta}},
	description={observed change},
}

\newglossaryentry{query}{type=symbols,
	sort={},
	name={\ensuremath{\vec{q}}},
	description={user query},
}
\newglossaryentry{tf_query}{type=symbols,
	sort={},
	name={\ensuremath{\Mat{T}^\mathrm{QW}}},
	description={pose of the query},
}
\newglossaryentry{occ_map}{type=symbols,
	sort={},
	name={\ensuremath{\mathbf{M}}},
	description={occupancy map},
}

%% file: sections/abstract.tex
\begin{abstract}
Robots deployed in real-world environments, such as homes, must not only navigate safely but also understand their surroundings and adapt to changes in the environment. To perform tasks efficiently, they must build and maintain a semantic map that accurately reflects the current state of the environment. Existing research on semantic exploration largely focuses on static scenes without persistent object-level instance tracking.
In this work, we propose an open-vocabulary, semantic exploration system for semi-static environments. Our system maintains a consistent map by building a probabilistic model of object instance stationarity, systematically tracking semi-static changes, and actively exploring areas that have not been visited for an extended period. In addition to active map maintenance, our approach leverages the map’s semantic richness with \ac{llm}-based reasoning for open-vocabulary object-goal navigation. This enables the robot to search more efficiently by prioritizing contextually relevant areas. 
We compare our approach against state-of-the-art baselines using publicly available object navigation and mapping datasets, and we further demonstrate real-world transferability in three real-world environments. Our approach outperforms the compared baselines in both success rate and search efficiency for object-navigation tasks and can more reliably handle changes in mapping semi-static environments. In real-world experiments, our system detects 95\% of map changes on average, improving efficiency by more than 29\% as compared to random and patrol strategies.

\end{abstract}

\glsresetall

%% file: sections/introduction.tex
\section{Introduction}

Humans can easily navigate unfamiliar environments using prior knowledge and adapt to changes, such as moved furniture or new objects.
For example, they might expect to find their reading glasses on a bedside table or near books.
Similarly, autonomous robots in everyday environments require semantic understanding, contextual reasoning, and adaptability to changing surroundings.

Object-goal navigation, where a robot must locate objects in unknown or partially known spaces, demands a tight integration of semantic understanding and spatial reasoning.
To address this challenge, both modular approaches combining localization, mapping, planning, and control with heuristic exploration strategies~\cite{yamauchi_frontier-based_1997,sun_frontiernet_2025} and learning-oriented methods~\cite{chaplot_object_2020, ye_auxiliary_2021} %
have been proposed in the past. More recently, hybrid methods have emerged to improve the exploration efficiency by more seamlessly embedding semantic understanding and reasoning into the pipeline~\cite{shah_foresightnav_2025,goel_semantically_2023,georgakis_learning_2022}. Yet most works still target static scenes, with any dynamic changes either masked as outliers or tracked only over consecutive frames. Many real-world changes, however, are semi-static (e.g., furniture or electronic devices can be shifted around) and are not directly observed~\cite{qian_pocd_2022,qian_pov-slam_2023}. In such scenarios, maintaining a consistent spatio-temporal environment representation is essential for long-term autonomy and the efficient execution of downstream tasks; this capability, however, remains underexplored in object-goal navigation tasks.
\begin{figure}
{
    \centering
    \footnotesize
   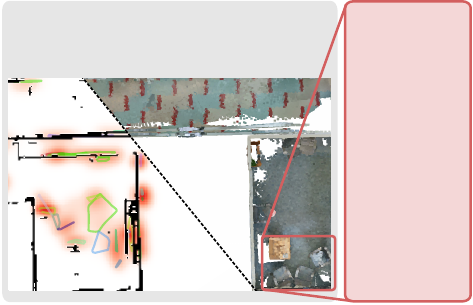
   \vspace{-0.3cm}
   \caption{An illustration of our proposed open-vocabulary semantic exploration approach for semi-static environments, where objects can be shifted, removed, or reintroduced. To account for such changes, the system explicitly maintains a stationarity score for each object instance and actively revisits regions of the map that are likely outdated. This enables the construction of an up-to-date metric-semantic map, which we use to prioritize contextually relevant areas during (unseen) object-goal navigation in semi-static scenes. \textcolor{\editc}{An overview of our work, including real-world experiments, can be found on our website \webpagelink{} and in our video \videolink{}.}%
   }
   \vspace{-0.3cm}
}
\end{figure}

\IEEEpubidadjcol
In this work, we propose a semantic exploration framework for semi-static environments (i.e., an environment where objects can be moved, removed, and/or (re)introduced). Our approach tightly couples mapping in semantic scenes with safe planning and control to leverage the efficiency of modular pipelines, while incorporating language-conditioned semantic understanding and reasoning to enable open-vocabulary, context-aware exploration. Our framework encompasses two modes of operation to support real-world applications: \textit{(i)} active map maintenance, in which the robot actively maintains an up-to-date metric-semantic map by revisiting regions of the map that are likely outdated while deprioritizing areas that are likely static, and \textit{(ii)} open-vocabulary semantic exploration, in which the robot performs object-goal navigation based on language inputs in unstructured, semi-static environments. Our contributions are summarized as follows:   
\begin{itemize}
    \item We propose a novel online open-vocabulary 3D mapping scheme that incorporates probabilistic change detection to track object instances under possible changes such as being moved, removed, or (re)introduced. %
    \item We introduce a semantic exploration method that allows a robot to actively revisit likely outdated map regions and efficiently perform object-goal navigation in semi-static environments.
    \item We show that our method outperforms state-of-the-art object-goal navigation and mapping baselines in changing environments through extensive evaluations on public datasets and further validate its efficacy in real-time, closed-loop experiments conducted across three real-world environments.
\end{itemize}

%% file: images/front_figure/front_figure.pdf_tex
\begingroup%
  \makeatletter%
  \providecommand\color[2][]{%
    \errmessage{(Inkscape) Color is used for the text in Inkscape, but the package 'color.sty' is not loaded}%
    \renewcommand\color[2][]{}%
  }%
  \providecommand\transparent[1]{%
    \errmessage{(Inkscape) Transparency is used (non-zero) for the text in Inkscape, but the package 'transparent.sty' is not loaded}%
    \renewcommand\transparent[1]{}%
  }%
  \providecommand\rotatebox[2]{#2}%
  \newcommand*\fsize{\dimexpr\f@size pt\relax}%
  \newcommand*\lineheight[1]{\fontsize{\fsize}{#1\fsize}\selectfont}%
  \ifx\svgwidth\undefined%
    \setlength{\unitlength}{226.77165354bp}%
    \ifx\svgscale\undefined%
      \relax%
    \else%
      \setlength{\unitlength}{\unitlength * \real{\svgscale}}%
    \fi%
  \else%
    \setlength{\unitlength}{\svgwidth}%
  \fi%
  \global\let\svgwidth\undefined%
  \global\let\svgscale\undefined%
  \makeatother%
  \begin{picture}(1,0.64375001)%
    \lineheight{1}%
    \setlength\tabcolsep{0pt}%
    \put(0,0){\includegraphics[width=\unitlength,page=1]{front_figure.pdf}}%
    \put(0.36177232,0.59){\color[rgb]{0,0,0}\makebox(0,0)[t]{\lineheight{1.25}\smash{\begin{tabular}[t]{c}Exploration Priority Map\\to \textbf{activly maintain} an up-to-date map\\and search objects\end{tabular}}}}%
    \put(0.86238503,0.61045464){\color[rgb]{0,0,0}\makebox(0,0)[t]{\lineheight{1.25}\smash{\begin{tabular}[t]{c}Initial\\Scene State\end{tabular}}}}%
    \put(0.86322046,0.19399009){\color[rgb]{0,0,0}\makebox(0,0)[t]{\lineheight{1.25}\smash{\begin{tabular}[t]{c}w/ Change\\Detection (Ours)\end{tabular}}}}%
    \put(0.86354775,0.40222234){\color[rgb]{0,0,0}\makebox(0,0)[t]{\lineheight{1.25}\smash{\begin{tabular}[t]{c}w/o Change\\Detection\end{tabular}}}}%
    \put(0,0){\includegraphics[width=\unitlength,page=2]{front_figure.pdf}}%
    \put(0.64695894,0.43795175){\color[rgb]{0,0,0}\rotatebox{0.17749406}{\makebox(0,0)[rt]{\lineheight{1.25}\smash{\begin{tabular}[t]{r}\scriptsize 3D Metric-Semantic Map\end{tabular}}}}}%
    \put(0.02706875,0.04327265){\color[rgb]{0,0,0}\rotatebox{-0.07824171}{\makebox(0,0)[lt]{\lineheight{1.25}\smash{\begin{tabular}[t]{l}\scriptsize Exploration Priority Map\end{tabular}}}}}%
    \put(0,0){\includegraphics[width=\unitlength,page=3]{front_figure.pdf}}%
  \end{picture}%
\endgroup%

%% file: sections/related_work.tex
\section{Related Work}

\begin{figure*}
{
   \footnotesize
   \centering
   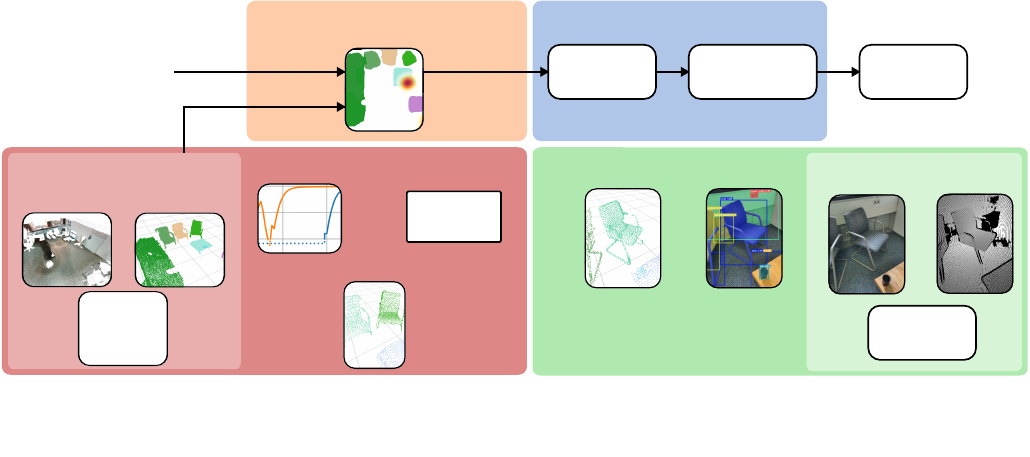
   \vspace{-1.7cm}
   \caption{Overview of our proposed system. We extract object candidates from the current pose and RGB-D frame (green). These are associated with objects in the semantic map, which is updated based on a probabilistic consistency estimate (red). Based on the scene belief, we build a semantic exploration priority map, indicating which map regions are relevant to current tasks -- maintaining an up-to-date map or object-goal navigation (orange). Finally, the robot leverages the priority map to select and navigate to sampled positions (blue).}\label{fig:block_diagram}
}
\vspace{-1.7em}
\end{figure*}

\subsection{Change Detection in Object-Aware Mapping}
While traditional mapping methods have focused on achieving high geometric precision, recent approaches have begun to incorporate semantic understanding to build object-level maps~\cite{grinvald2019volumetric,
hughes2022hydra}, thereby enabling more advanced contextual reasoning and decision-making~\cite{gu_conceptgraphs_2023}. A common assumption in localization and mapping systems, however, is that the environment is static, which limits real-world applicability.

Several efforts have emerged to relax this assumption. One common strategy is to identify moving objects and mask them as outliers~\cite{yu_ds-slam_2018, 
schmid_dynablox_2023}. While this can reduce artifacts in the environment map, it also leads to information loss and instability when large portions of the scene are dynamic~\cite{tipaldi_lifelong_2013}. Another approach is to jointly track the camera pose and dynamic components~\cite{xu2019mid
}, but such methods often require changes to be observed across consecutive frames. This is limiting in real-world settings, where environment changes are often semi-static, meaning the changes are observed only at discrete times (e.g., when a robot revisits a scene).

A few recent works have attempted to address this challenge.
Khronos~\cite{schmid_khronos_2024}, for example, constructs metric-semantic maps that capture both short- and long-term changes through fragment consistency. However, it associates observations with the map solely based on geometric data, which prevents the tracking of semi-static objects at the instance level. Panoptic TSDF~\cite{schmid_panoptic_2022} enforces object-level consistency, but its voxel-change counting makes it vulnerable to sensor noise and localization errors. In our work, to maintain a spatio-temporally consistent map, we instead incorporate a probabilistic change detection formulation~\cite{qian_pocd_2022,qian_pov-slam_2023} that fuses semantic prior and geometric consistency to robustly track object instances, even when they disappear and later reappear elsewhere. Moreover, beyond passively bookkeeping changes, our framework closes the perception-action loop to actively improve the map over likely non-stationary regions and consequently enables more efficient completion of downstream tasks.

\subsection{Semantic Exploration}

Semantic exploration refers to the task of using semantic cues to efficiently explore an unknown (or changed) environment \cite{gervet_navigating_2023} to support, for instance, mapping.
Non-semantic exploration relies on heuristics to select frontiers~\cite{yamauchi_frontier-based_1997,papatheodorou_finding_2023}, while recent semantic approaches guide navigation toward areas likely to reveal relevant space, such as doors~\cite{sun_frontiernet_2025}. Generative models further extend this by predicting unseen areas to direct exploration~\cite{shah_foresightnav_2025}.

Another line of work deals with object-goal navigation, where the robot is tasked with finding an unseen object of a specific category. Semantic priors are often learned via \ac{rl}%
~\cite{chaplot_object_2020, ye_auxiliary_2021},
but end-to-end methods lack modularity \cite{gervet_navigating_2023}. This is addressed in \cite{goel_semantically_2023, georgakis_learning_2022}, which proposes learning top-down semantic maps for use in classical control schemes such as \ac{mpc} \cite{goel_semantically_2023}. However, these methods have several limitations. They are constrained by their training data, often rely on prebuilt maps \cite{chaplot_object_2020, goel_semantically_2023}, and lack long-term memory of the space explored%
~\cite{ye_auxiliary_2021, georgakis_learning_2022, yokoyama2024vlfm}. 
Moreover, all these approaches assume static environments. While methods with short-term memory select frontiers at map edges, we can select internal frontiers to account for potential scene changes.

The open-vocabulary paradigm removes the need for a predefined training-time object set~\cite{gu_conceptgraphs_2023}, which is made possible by fusing the output of 2D foundation models with 3D information. In open-vocabulary object-goal navigation, the exploration frontiers are selected by, for instance, ranking observed objects using their CLIP embedding similarity to the target~\cite{jiang_dualmap_2025, laina_findanything_2025}, or by using \acp{llm} directly to rank frontiers by object descriptions~\cite{zhou_esc_2023, dorbala_can_2024}. OK-Robot~\cite{liu_ok-robot_2024} and ConceptGraphs~\cite{gu_conceptgraphs_2023} are closely related to this work, but they cannot deal with changing environments. DynaMem~\cite{liu2024dynamem} allows updates to its map while also supporting open-vocabulary navigation. However, it can not leverage semantic information to explore towards unseen object goals.

In this work, we propose task-specific open-vocabulary exploration that combines (unseen) object-goal navigation with maintaining an up-to-date map in semi-static scenes. We leverage priors from an \ac{llm} and a change detection framework to build top-down semantic maps that guide exploration toward areas of interest--those likely outdated or semantically relevant to the object-goal.%

%% file: figures/new_block_diagram_v2.pdf_tex
\begingroup%
  \makeatletter%
  \providecommand\color[2][]{%
    \errmessage{(Inkscape) Color is used for the text in Inkscape, but the package 'color.sty' is not loaded}%
    \renewcommand\color[2][]{}%
  }%
  \providecommand\transparent[1]{%
    \errmessage{(Inkscape) Transparency is used (non-zero) for the text in Inkscape, but the package 'transparent.sty' is not loaded}%
    \renewcommand\transparent[1]{}%
  }%
  \providecommand\rotatebox[2]{#2}%
  \newcommand*\fsize{\dimexpr\f@size pt\relax}%
  \newcommand*\lineheight[1]{\fontsize{\fsize}{#1\fsize}\selectfont}%
  \ifx\svgwidth\undefined%
    \setlength{\unitlength}{493.22834646bp}%
    \ifx\svgscale\undefined%
      \relax%
    \else%
      \setlength{\unitlength}{\unitlength * \real{\svgscale}}%
    \fi%
  \else%
    \setlength{\unitlength}{\svgwidth}%
  \fi%
  \global\let\svgwidth\undefined%
  \global\let\svgscale\undefined%
  \makeatother%
  \begin{picture}(1,0.44252874)%
    \lineheight{1}%
    \setlength\tabcolsep{0pt}%
    \put(0,0){\includegraphics[width=\unitlength,page=1]{new_block_diagram_v2.pdf}}%
    \put(0.37810772,0.42467079){\color[rgb]{0,0,0}\makebox(0,0)[t]{\lineheight{1.25}\smash{\begin{tabular}[t]{c}Generate Top-Down Exploration\\Priority Map $f_{\mathrm{task}}(\cdot \mathop{|} \gls{obj_lib}_t , \gls{query})$\end{tabular}}}}%
    \put(0.86752416,0.36728033){\color[rgb]{0,0,0}\makebox(0,0)[lt]{\lineheight{1.25}\smash{\begin{tabular}[t]{l}Robot\end{tabular}}}}%
    \put(0.67527036,0.37908873){\color[rgb]{0,0,0}\makebox(0,0)[lt]{\lineheight{1.25}\smash{\begin{tabular}[t]{l}Global Path\\Planning \& MPC\end{tabular}}}}%
    \put(0.54351405,0.36892765){\color[rgb]{0,0,0}\makebox(0,0)[lt]{\lineheight{1.25}\smash{\begin{tabular}[t]{l}Sample Goal \end{tabular}}}}%
    \put(0.72330267,0.14823583){\color[rgb]{0,0,0}\makebox(0,0)[t]{\lineheight{1.25}\smash{\begin{tabular}[t]{c}Open-Vocabulary\\Segmentation\end{tabular}}}}%
    \put(0.29063704,0.28608174){\color[rgb]{0,0,0}\makebox(0,0)[t]{\lineheight{1.25}\smash{\begin{tabular}[t]{c}Update\\Stationarity\end{tabular}}}}%
    \put(0.50303721,0.27793809){\color[rgb]{0,0,0}\makebox(0,0)[rt]{\lineheight{1.25}\smash{\begin{tabular}[t]{r}Sec. \ref{sec:object_detection_and_updating}\end{tabular}}}}%
    \put(0.52551179,0.27793809){\color[rgb]{0,0,0}\makebox(0,0)[lt]{\lineheight{1.25}\smash{\begin{tabular}[t]{l}Sec. \ref{sec:observations}\end{tabular}}}}%
    \put(0.52494479,0.31460836){\color[rgb]{0,0,0}\makebox(0,0)[lt]{\lineheight{1.25}\smash{\begin{tabular}[t]{l}Sec.~\ref{sec:planning_and_control}\end{tabular}}}}%
    \put(0.50295197,0.31456085){\color[rgb]{0,0,0}\makebox(0,0)[rt]{\lineheight{1.25}\smash{\begin{tabular}[t]{r}Sec. \ref{sec:heatmap}\end{tabular}}}}%
    \put(0.02254137,0.38895678){\color[rgb]{0,0,0}\makebox(0,0)[lt]{\lineheight{1.25}\smash{\begin{tabular}[t]{l}User query $\gls{query}$:\\"Find my book!"\\"Maintain the map!"\end{tabular}}}}%
    \put(0.11326674,0.28081777){\color[rgb]{0,0,0}\makebox(0,0)[t]{\lineheight{1.25}\smash{\begin{tabular}[t]{c}Semantic Map\end{tabular}}}}%
    \put(0.17570822,0.26034364){\color[rgb]{0,0,0}\makebox(0,0)[t]{\lineheight{1.25}\smash{\begin{tabular}[t]{c}Object\\Library $\gls{obj_lib}_t$\end{tabular}}}}%
    \put(0.06569549,0.26296294){\color[rgb]{0,0,0}\makebox(0,0)[t]{\lineheight{1.25}\smash{\begin{tabular}[t]{c}Background\\$\Mat{P}_{\text{bg},t}$\end{tabular}}}}%
    \put(0.11914619,0.14236433){\color[rgb]{0,0,0}\makebox(0,0)[t]{\lineheight{1.25}\smash{\begin{tabular}[t]{c}Missing\\Objects\\$\gls{obj_lib}_{\mathrm{mis},t}$ \end{tabular}}}}%
    \put(0.89722469,0.12540035){\color[rgb]{0,0,0}\makebox(0,0)[t]{\lineheight{1.25}\smash{\begin{tabular}[t]{c}Pose $\gls{tf_cam}_t$\\(VIO)\end{tabular}}}}%
    \put(0.36680345,0.19324859){\color[rgb]{0,0,0}\makebox(0,0)[t]{\lineheight{1.25}\smash{\begin{tabular}[t]{c}Expected\\Objects $\gls{obj_lib}_{t,\mathrm{exp}}$\end{tabular}}}}%
    \put(0.44076428,0.23719192){\color[rgb]{0,0,0}\makebox(0,0)[t]{\lineheight{1.25}\smash{\begin{tabular}[t]{c}Compare \&\\Associate\end{tabular}}}}%
    \put(0.6054489,0.14710043){\color[rgb]{0,0,0}\makebox(0,0)[t]{\lineheight{1.25}\smash{\begin{tabular}[t]{c}Object\\Canditates $\gls{detections}_t$\end{tabular}}}}%
    \put(0.88411234,0.28020796){\color[rgb]{0,0,0}\makebox(0,0)[t]{\lineheight{1.25}\smash{\begin{tabular}[t]{c}Posed RGB-D\\Frame $\gls{frame}_t$\end{tabular}}}}%
    \put(0,0){\includegraphics[width=\unitlength,page=2]{new_block_diagram_v2.pdf}}%
  \end{picture}%
\endgroup%

%% file: sections/methodology.tex
\section{Problem Formulation}

Similar to~\cite{gu_conceptgraphs_2023, liu_ok-robot_2024, liu2024dynamem}, our system takes RGB-D frames~$\gls{frame}_t$ along with camera poses~$\gls{tf_cam}[_t]$, which are assumed known and obtained here via a \ac{vio} system~\cite{seiskari_hybvio_2022}. It incrementally builds and updates a semantic map of a semi-static environment comprising an object library~$\gls{obj_lib}_t$, a missing object library~$\gls{obj_lib}_{\mathrm{mis}, t}$, and a background point cloud~$\Mat{P}_{\text{bg},t}$, all of which can be initialized as empty sets or based on a prior map (if available). Given a user query~$\gls{query}$~(e.g., ``Find my glasses!'' or ``Maintain the map!''), we generate an exploration priority map which allows the robot to either search for the requested object as efficiently as possible or actively revisit likely outdated map regions to maintain a spatial-temporal consistent 3D environment representation.

We note that, similar to other open-vocabulary object-navigation works such as~\cite{liu2024dynamem}, our work focuses on wheeled robots moving on a 2D floor plane. Thus, while we maintain a full 3D map of the environment, a 2D priority map is used for semantic exploration. %
However, since a 3D map is readily available, our approach can be naturally extended to settings that require full 3D spatial reasoning. %

\section{Methodology}
An overview of our method is presented in Fig.~\ref{fig:block_diagram}. At each timestep~$t$, we process an RGB-D frame~$\gls{frame}_t$ with its camera pose~$\gls{tf_cam}[_t]$ to detect a set of currently visible object candidates~$\gls{detections}_t$; each object candidate~$\gls{detection}_{t,j}=(\gls{det_tf}[_j], \gls{det_pc}_j, \gls{det_f}_j, \gls{det_c}_j)$ comprises, a 6-\ac{dof} pose $\gls{det_tf}[_j]$, a point cloud $\gls{det_pc}_j$, a visual feature $\gls{det_f}_j$, and an open-vocabulary class~$\gls{det_c}_j$. 

The object candidates are used to update the semantic map, comprising the object library~$\gls{obj_lib}_t$, missing object library~$\gls{obj_lib}_{\mathrm{mis}, t}$ containing objects that have been explicitly observed to vanish from the scene, and the accumulated background point cloud~$\Mat{P}_{\text{bg},t}$ representing static structures. The object library~$\gls{obj_lib}_t$ and missing object library~$\gls{obj_lib}_{\mathrm{mis}, t}$ are two disjoint sets of objects; each object $\gls{obj}_i = (\gls{obj_tf}[_i], \gls{obj_pc}_i, \gls{obj_b}_i, \gls{obj_f}_i, \gls{obj_c}_i, \gls{obj_s}_i, (\gls{obj_t_first}[_i], \gls{obj_t_disappear}[_i]), \theta_i)$ comprises, a 6-DOF pose $\gls{obj_tf}[_i]$, a point cloud $\gls{obj_pc}_i$, a visual feature $\gls{obj_f}_i$, an open-vocabulary class $\gls{obj_c}_i$, a prior stationary label $\gls{obj_s}_i \in \{\texttt{static}, \texttt{dynamic}\}$ describing whether objects from class $\gls{obj_c}_i$ are typically moved around or not~(obtained from an \ac{llm}), a first observed time $\gls{obj_t_first}[_i]$ and latest vanishing time $\gls{obj_t_disappear}_i$, and distribution parameters $\theta_i$ describing the current belief of the geometric change $\gls{geom_change}_i$ and stationary score $\gls{pocd_score}_i$, which inform map updates and maintenance. Given a query~$\gls{query}$ (e.g., ``Find my book!'' or ``Maintain the map!'') and scene belief~$\gls{obj_lib}_t$, we build an exploration priority map~$f_{\mathrm{task}}(\cdot \mathop{|} \gls{obj_lib}_t , \gls{query})$ to sample target waypoints~$\vec{w}^*$, executed via a global planner and \ac{mpc} for safe navigation.  Each component is further detailed in the respective subsections below.

\subsection{Current View Object Candidates}\label{sec:observations}

Current object candidates~$\gls{detections}_t = \{ \gls{detection}_{t,j} \}_{j=1,...,J}$ are extracted from the RGB-D frame~$\gls{frame}_t$ and its pose~$\gls{tf_cam}_t$~\cite{gu_conceptgraphs_2023}. %
In particular, the RGB image is segmented using SAM~\cite{Kirillov_2023_ICCV} into binary object masks~$\mathcal{M}_t = \{ \Mat{M}_{t,j} \}_{j=1,...,J}$. Using each mask~$\Mat{M}_{t,j}$ a visual feature vector~$\gls{det_f}_j$ and class label~$\gls{det_c}_j$ are computed with CLIP~\cite{radford2021learning}, and the RGB-D frame is segmented into point clouds~$\gls{det_pc}_j$, yielding the object candidate $\gls{detection}_{t,j}$.
Additionally, we filter out observations beyond a maximum distance $d_{\mathrm{max}}$ to the camera. The background point cloud~$\Mat{P}_{\text{bg},t}$ is updated with points not covered by any object mask (i.e., points within~$\neg \bigcup_{j=1}^{J} \Mat{M}_{t,j}$).

\subsection{Scene Belief Update}\label{sec:object_detection_and_updating}

The object libraries~$\gls{obj_lib}_{t-1}$ and~$\gls{obj_lib}_{\mathrm{mis},t-1}$ are updated by associating the current object candidates~$\gls{detections}_t$ with objects expected to be visible in the current camera view and by adding unmatched candidates as new objects. The stationarity scores of all objects are then updated accordingly.

To handle objects that may temporarily disappear or change position across frames (e.g., a chair disappearing and reappearing later elsewhere), object-to-object association is performed within~$\gls{obj_lib}_t$, allowing corresponding~$\gls{obj}_i, \gls{obj}_j \in \gls{obj_lib}_t$ to be merged. To avoid exhaustive pairwise comparisons, candidate pairs are selected based on their stationarity scores. These steps are detailed below.

\subsubsection{Expected-View Association}
At this stage, we restrict the matching process to objects that are expected to be visible within the current camera's field of view. The goal is to obtain a map from the current object candidates to the expected objects~$\gls{detections}_t \mapsto \gls{obj_lib}_{t,\mathrm{exp}} \cup \{ \mathrm{None} \}$.

The subset of expected objects~$\gls{obj_lib}_{t,\mathrm{exp}} \subseteq \gls{obj_lib}_t$ is determined based on the current camera pose~$\gls{tf_cam}[_t]$ and intrinsics. 
For each object~$\gls{obj}_i \in \gls{obj_lib}_{t-1}$, we filter out points in its point cloud~$\gls{obj_pc}_i$ that exceed the maximum distance~$d_{\mathrm{max}}$ from the camera, and project the remaining points onto the image plane. This yields the subset~$\gls{obj_pc}_{\mathrm{vis}, i} \subseteq \gls{obj_pc}_i$, representing the object part seen from the current view; the object is considered expected if its visible point fraction exceeds a given threshold, yielding
    $\gls{obj_lib}_{t,\mathrm{exp}} = \left\{ \gls{obj}_i \in \gls{obj_lib}_{t-1} \middle| \frac{|\gls{obj_pc}_{\mathrm{vis}, i}|}{|\gls{obj_pc}_i|} \geq \tau_{\mathrm{expected}} \right\}$ \;.

The matching process relies on two similarity measures:
\begin{itemize}
    \item \textbf{Semantic similarity}~$S_{\mathrm{sem}} \mathop{:} \gls{detections} \times \gls{obj_lib} \to [0,1]$, defined as the cosine similarity between the candidate's and the object's visual feature vectors~$\gls{det_f}_j$ and~$\gls{obj_f}_i$:
    \begin{equation}\label{eq:similarity_sem}
        S_{\mathrm{sem}}(\gls{detection}_j, \gls{obj}_i) = \gls{det_f}_j^T \gls{obj_f}_i / (\lVert \gls{det_f}_j \rVert_2 \lVert \gls{obj_f}_i \rVert_2) \;.
    \end{equation}%
    \item \textbf{Geometric similarity}~$S_{\mathrm{geo}} \mathop{:} \gls{detections} \times \gls{obj_lib} \to [0,1]$, computed as the fraction of points in the candidate's point cloud~$\gls{det_pc}_j$ whose nearest neighbor in the object's visible point cloud~$\gls{obj_pc}_{i,\mathrm{vis}}$ is within distance~$d_{\mathrm{voxel\,size}}$. With~$\gls{det_pc}_{j} \cap \gls{obj_pc}_{i,\mathrm{vis}}$ denoting this subset:
    \begin{equation}\label{eq:similarity_geo}
        \hspace{-2em}S_{\mathrm{geo}}(\gls{detection}_j, \gls{obj}_i) = | \gls{det_pc}_{j} \cap \gls{obj_pc}_{i, \mathrm{vis}} | / \min (|\gls{det_pc}_{j}|, |\gls{obj_pc}_{i, \mathrm{vis}}|) \;.
    \end{equation}

\end{itemize}

Matching candidates~$\gls{detections}_t$ to expected objects~$\gls{obj_lib}_{t,\mathrm{exp}}$~(Alg.~\ref{alg:det_to_obj_association}) first assumes objects are stationary (first \texttt{for}-loop), greedily pairing them by geometric similarity, with matches accepted only if both geometric and semantic scores exceed the thresholds~$\tau_{\mathrm{geo}}$ and ~$\tau_{\mathrm{sem}}$, respectively. The second step (second \texttt{for}-loop) matches remaining candidates that might have been moved by prioritizing semantic similarity and verifying alignment via \ac{icp} rather than the geometric measure $S_{\mathrm{geo}}(\cdot, \cdot)$. %
Matches require semantic similarity above~$\tau_{\mathrm{sem}}$ and ICP \ac{rmse} below~$d_{\mathrm{ICP}}$. This defers costly \ac{icp} until necessary. Finally, matched candidates are merged with their corresponding objects in~$\gls{obj_lib}_t$, and unmatched ones are added as new objects.
\begin{algorithm}[b]
\caption{2-Step Candidate to Object Association}\label{alg:det_to_obj_association}
    \coloredited
    \begin{algorithmic}
    \State \textbf{Input:} $\gls{detections}_t = \{ \gls{detection}_{t,j} \}_{j=1,...,J}$, $\gls{obj_lib}_{t,\mathrm{exp}} = \{ \gls{obj}_i \}_{i=1,...,I}$
    \State $\mathcal{Y}_{\mathrm{matches}} \gets \varnothing $
    \ForAll{$\gls{detection}_j \in \gls{detections}_t$}     \Comment{Match \coloredited assuming stationary objects}
        \State $\hat{\Mat{o}} \gets \mathrm{argmax}_{\Mat{o} \in \gls{obj_lib}_{t,\mathrm{exp}}} \; S_{\mathrm{geo}}(\gls{detection}_j, \Mat{o})$
        \If{$S_{\mathrm{geo}}(\gls{detection}_j, \hat{\Mat{o}}) > \tau_{\mathrm{geo}} \land S_{\mathrm{sem}}(\gls{detection}_j, \hat{\Mat{o}}) > \tau_{\mathrm{sem}}$}
            \State $\mathcal{Y}_{\mathrm{matches}} \gets \mathcal{Y}_{\mathrm{matches}} \cup \{ \gls{detection}_j \mapsto \hat{\Mat{o}} \}$
        \EndIf
    \EndFor
    \LComment{Match \coloredited without assuming stationary objects}
    \ForAll{$\gls{detection}_j  \in  \gls{detections}_t \setminus \operatorname{dom}(\gls{detections}_{\mathrm{matches}})$} 
        \State $\hat{\Mat{o}} \gets \mathrm{argmax}_{\Mat{o} \in \gls{obj_lib}_{t,\mathrm{exp}} \setminus \operatorname{im}(\gls{detections}_{\mathrm{matches}})} \; S_{\mathrm{sem}}(\gls{detection}_j, \Mat{o})$
        \If{$S_{\mathrm{sem}}(\gls{detection}_j, \hat{\Mat{o}}) > \tau_{\mathrm{sem}}$}
            \If{$\mathrm{RMSE}_{\mathrm{ICP}}(\gls{det_pc}_j, \gls{obj_pc}_i) \leq d_{\mathrm{ICP}}$}
                \State $\mathcal{Y}_{\mathrm{matches}} \gets \mathcal{Y}_{\mathrm{matches}} \cup \{ \gls{detection}_j \mapsto \hat{\Mat{o}} \}$
            \EndIf
        \EndIf
    \EndFor
    \State \Return $\mathcal{Y}_{\mathrm{matches}} \cup \{ \gls{detection} \mapsto \mathrm{None} \mid \gls{detection} \in \gls{detections}_t \setminus \operatorname{dom}(\gls{detections}_{\mathrm{matches}}) \}$
    \end{algorithmic}
\end{algorithm}

\subsubsection{Stationarity Score Update}

We maintain a probabilistic stationarity model for every object instance based on~\cite{qian_pocd_2022}, which grounds our decision for when to translate or delete objects in our believed scene representation. Each object~$\gls{obj}_i$ contains a belief of its geometric change $\gls{geom_change}_i \in \mathbb{R}$ and stationarity score $\gls{pocd_score}_i \in [0,1]$ (i.e., $p(\gls{pocd_score}_i, \gls{geom_change}_i | \dots )$, conditioned as in \eqref{eq:pocd_update}). Stationarity score~$\gls{pocd_score}_i$ represents the likelihood that the object~$\gls{obj}_i$ is still located where last observed, the geometric change~$\gls{geom_change}_i$ measures how much the object has moved since the last timestep (in our case simply the distance its point cloud center of mass has moved).

At each timestep, we compute the measured change $\gls{change}_{t,i} \in \mathbb{R}$ of each object (i.e., the distance the object has translated since the last timestep). Based on this change~$\gls{change}_{t,i}$ and the semantic stationarity prior~$\gls{obj_s}_i$, the belief is updated each step in a Bayesian fashion \cite{qian_pocd_2022}(we omit the index $i$ here):
\begin{equation}\label{eq:pocd_update}%
    p(\gls{pocd_score}, \gls{geom_change} | \gls{change}_{1:t}, \gls{obj_s}) \propto 
    p(\gls{change}_{t} | \gls{pocd_score}, \gls{geom_change}, \gls{change}_{1:t-1} ) 
    p(\gls{pocd_score}, \gls{geom_change} | \gls{change}_{1:t-1}, \gls{obj_s}) \;.
\end{equation}
Later, each object's expected stationarity score~$\mathbb{E}[\gls{pocd_score}_i]$ informs future updates and map maintenance.

\begin{figure}
    \centering
    \includegraphics[clip, trim={0.01cm, 0.9cm, 0, 1.15cm}]{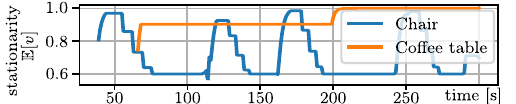}
    \caption{Different decaying of the stationarity score $\mathbb{E}[v_i]$ for an object with $\gls{obj_s}_i = \texttt{dynamic}$ (Chair) and an object with $\gls{obj_s}_i = \texttt{static}$ (Coffee table). The objects are observed only at the times their stationary score increases; otherwise, they are not in the robot's view.}
    \label{fig:stationarity_dynamic_vs_static}
\end{figure}

We can directly measure the change~$\gls{change}_{t,i}$ only for objects that have been reobserved. To model the growing uncertainty of out-of-view objects, we inject a synthetic change to~$\gls{change}_{t,i}$ with magnitude defined based on the semantic prior $\gls{obj_s}_i$. For a likely \texttt{static}/\texttt{dynamic} object, the magnitude is correspondingly smaller/larger; this will correspondingly result in a slower/faster decay in the stationarity score of the object instance.
To prevent premature deletion from the scene belief, we stop the decay process once the expected stationarity score~$\mathbb{E}[\gls{pocd_score}_i]$ reaches the removal threshold~$\theta_{\mathrm{r}}$. Figure~\ref{fig:stationarity_dynamic_vs_static} illustrates the evolution of~$\mathbb{E}[\gls{pocd_score}_i]$ for a \texttt{dynamic} object (chair) and a \texttt{static} object (coffee table). The dynamic object is assumed to move more frequently, resulting in a faster decay of its stationarity score.

\subsubsection{Object-to-Object Association: Removal, Reintroduction, and Transformation}

In addition to immediate object detection, we must handle objects that disappear from view and may later reappear, potentially at a different location. 
Removal and translation decisions depend on each object's stationarity~$\mathbb{E}[\gls{pocd_score}_i]$, evaluated against the removal and translation thresholds, $0 \leq \theta_{\mathrm{r}} < \theta_{\mathrm{t}} \leq 1$.
These thresholds balance system responsiveness and robustness to perception noise.

Objects with $\mathbb{E}[\gls{pocd_score}_i] \leq \theta_{\mathrm{r}}$ are moved from the active object library~$\gls{obj_lib}_t$ to the missing object library~$\gls{obj_lib}_{\mathrm{mis},t}$, where they are kept for potential future reidentification.

We consider two groups of objects for reidentification: objects in~$\gls{obj_lib}_t$ with~$\theta_{\mathrm{r}} < \mathbb{E}[\gls{pocd_score}_i] \leq \theta_{\mathrm{t}}$, and objects in the missing library~$\gls{obj_lib}_{\mathrm{mis}, t}$. For each object~$\gls{obj}_k$ from these groups that disappeared at time~$\gls{obj_t_disappear}[_k]$, we search for matching objects which appeared in a temporal window $\pm \overline{\tau}$ around its disappearance $\mathcal{O}_{t \approx \gls{obj_t_disappear}[_k]} = \left\{ \gls{obj}_i \in \gls{obj_lib}_{t} \;\middle|\;  |\gls{obj_t_first}[_i]-\gls{obj_t_disappear}[_k]| \leq \overline{\tau} \right\}$.    
Matching is performed between~$\gls{obj}_k$ and candidates in~$\mathcal{O}_{t \approx \gls{obj_t_disappear}_k}$ using \textcolor{\editc}{the same semantically conditioned \ac{icp} as in the second \texttt{for}-loop of Alg.~\ref{alg:det_to_obj_association}}. If~$\gls{obj}_k$ matches~$\gls{obj}_i \in \mathcal{O}_{t \approx \gls{obj_t_disappear}_k}$, the two are merged. If the matched counterpart was previously marked missing (\(\gls{obj}_i \in \gls{obj_lib}_{\mathrm{mis},t}\)), it is moved from~$\gls{obj_lib}_{\mathrm{mis},t}$ to the active library~$\gls{obj_lib}_t$ before merging. This enables us to differentiate between objects of the same class and to reintegrate previously observed objects when they reappear~(see results in \textcolor{\editc}{Fig.~\ref{fig:reidentificaion}}).

\subsection{Exploration Priority Map}\label{sec:heatmap}

Our navigation framework is based on an exploration priority map, which represents the likelihood of completing the current task $\gls{query}$, either object goal navigation or map maintenance, at each 2D point $\vec{x} \in \mathbb{R}^2$ in the environment. It is defined as $f_{\mathrm{task}}(\vec{x} \mathop{|} \gls{obj_lib}_t , \gls{query}) \mathop{:} \mathbb{R}^2 \mapsto [0,\infty)$, with $\int_{\mathbb{R}^2} f_{\mathrm{task}}(\vec{x} \mathop{|} \gls{obj_lib}_t , \gls{query}) \mathrm{d}\vec{x}=1$, conditioned on the current objects $\gls{obj_lib}_t$ and on the query $\gls{query}$.

We compute the exploration priority map as a superposition of per-object maps $f(\vec{x} \mathop{|} \gls{obj}) : \mathbb{R}^2 \to [0,\infty)$, weighted by task-dependent relevancy scores $\lambda(\gls{obj} \mid \gls{query}) \in [0,1]$, yielding 
    $f_{\mathrm{task}}(\vec{x} \mathop{|} \gls{obj_lib}_t , \gls{query}) \propto \sum_{\gls{obj} \in \gls{obj_lib}_t} \lambda(\gls{obj} \mathop{|} \gls{query}) f(\vec{x} \mathop{|} \gls{obj} )$ .

\subsubsection{Per-Object Exploration Priority Map}

The per-object exploration priority map $f(\vec{x} \mathop{|} \gls{obj}_i)$ is derived by smoothing the object's occupancy shadow with a Gaussian kernel
    $f(\vec{x} \mid \gls{obj}_i) \propto B_i(\vec{x}) * K\big(\vec{x} \mid 0, \sigma(\mathbb{E}[v_i])^2\big)$ with $\int_{\mathbb{R}^2} f(\vec{x} \mid \gls{obj}_i) \, d\vec{x} = 1$,  
where $B_i(\vec{x})$ is a binary mask indicating the ground-plane region occupied (or shadowed) by object $\gls{obj}_i$.
The Gaussian kernel $K(\vec{x} \mathop{|} 0, \sigma(\mathbb{E}[v_i])^2)$ has zero mean and a standard deviation $\sigma(\mathbb{E}[v_i])$ dependent on the object's stationarity $\mathbb{E}[v_i]$.
This smoothing expands the exploration priority map for objects with lower stationarity, reflecting higher uncertainty over their position. We choose the standard deviation $\sigma$ as $\sigma(v) = \frac{ v^{-1} - 1 }{ v_{\text{search}}^{-1} - 1 } (r_{\text{search}} - \sigma_{\text{measure}}) + \sigma_{\text{measure}}$.
This models three uncertainty cases: \textit{(i)} as stationarity $v$ approaches $0$, uncertainty becomes unbounded (no prior knowledge of location); \textit{(ii)} at an intermediate stationarity~$v_{\text{search}}$, a tunable search radius~$r_{\text{search}}$ applies; and \textit{(iii)} as~$v$ approaches $1$, uncertainty is dominated by measurement noise $\sigma_{\text{measure}}$.

\subsubsection{Object Semantic Relevancy}

Each object~$\gls{obj}_i$ is scored with a relevancy~$\lambda(\gls{obj}_i \mathop{|} \gls{query}) \in [0,1]$ that reflects its importance for the current task. This score is computed differently depending on whether the query \gls{query} is the map maintenance task or an object search task.

For map maintenance, the goal is to build an accurate map by prioritizing regions likely to be outdated. Relevancy is based on the object stationarity~$\mathbb{E}[v_i]$ that captures both perception confidence and prior stationarity class $\lambda(\gls{obj}_i \mathop{|} \gls{query} = \mathrm{maintenance}) = \frac{f_{\mathrm{beta}}(\mathbb{E}[v_i]; \alpha, \beta)}{\max_v f_{\mathrm{beta}}(v; \alpha, \beta)}$ ,
where $f_{\mathrm{beta}}$ is the Beta PDF with parameters $(\alpha,\beta)=(5,6)$ %
The Beta distribution is a design choice for mapping stationarity to relevancy: It allows flexible shaping to emphasize low-stationarity objects without overemphasizing highly dynamic ones, but other options are possible.

In object search tasks, the robot locates a target object specified by a language query $\gls{query}$. An \ac{llm} estimates the likelihood of the target being successful near each known object class $\gls{obj_c}_i$ by answering a prompt like “How likely is the query $\gls{query}$ successful near a $\gls{obj_c}_i$?
Return the answer as an integer between 0 and 100, and append a sentence to justify your estimation.” separately for each query-class pair $(\gls{query}, \gls{obj_c}_i)$ pair.
\textcolor{\editc}{The returned value serves directly as the relevancy $\lambda(\gls{obj}_i \mid \gls{query})$, guiding search toward semantically related areas, even for unseen objects.}
For instance, ``Where is my plate?'' yields high relevance for tableware, tables, or cabinets (see Fig.~\ref{fig:search_plate}).

\subsection{Planning and Control}\label{sec:planning_and_control}

\begin{figure}
    \centering
    \vspace{-0.5mm} %
    \begin{tikzpicture}
        \node {\includegraphics[clip, trim={0, 4.9mm, 0, 4mm}]{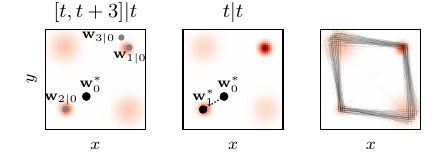}};
        \node [rectangle, fill=white, minimum width=4mm, minimum height=4mm] at (-3.25cm, 0) {};
    \end{tikzpicture}
    \vspace{-0.5cm}
    \caption{Illustration of how we sample from the exploration priority map. Last reached target waypoint $\vec{w}^*_{0}$ and iteratively sampled candidate waypoints $\vec{w}_{[1,3]|0}$ ($M=3$) (left). The closest candidate becomes the next target waypoint $\vec{w}^*_{1}$ (middle). Trajectory $\vec{w}^*_{[0,750]}$ produced after applying this sampling strategy for $750$ steps (right).}\label{fig:ergodic_sampling}
\end{figure}
To this end we want to mimic the priority map $f_{\mathrm{task}}(\vec{x} \mathop{|} \gls{obj_lib}_t , \gls{query})$ with the robot's infinite-time trajectory $\vec{x}_{[0,\infty)}$. While this control problem naturally suggests an ergodic control approach \cite{%
low_drozbot_2022}, we adopt a sampling-based strategy that enables global path planning, which mitigates the problem of local minima. We navigate to new target waypoints $\vec{w}^* \in \mathbb{R}^2$ until the task~$\gls{query}$ is complete or updated. Navigation is done by planning a collision-free path with $A^*$ and tracking it via \ac{mpc}.

To illustrate waypoint selection, assume the $i$-th waypoint~$\vec{w}^*_i$ was reached at time $t$. Then, the goal is to select the next target waypoint~$\vec{w}^*_{i+1}$ given the sequence of all past waypoints~$\vec{w}^*_{[0,i]}$ such that their low-passed density estimate $\hat{f}(\vec{x} \mathop{|} \vec{w}^*_{[0,i+1]})$ aligns with the exploration priority map $f_{\mathrm{task}}(\vec{x} \mathop{|} \gls{obj_lib}_t, \gls{query})$. For this, we could sample the next waypoint~$\vec{w}^*_{i+1}$ directly from the error distribution $e_{i|i}(\vec{x}) = f_{\mathrm{task}}(\vec{x} \mathop{|} \gls{obj_lib}_t, \gls{query}) - \hat{f}(\vec{x} \mathop{|} \vec{w}^*_{[0,i]})$,
which, however, would produce oscillation between modes of the exploration map. Instead, we iteratively sample $M$ candidate waypoints $\vec{w}_{[i+1,i+M]|i}$ and select the closest candidate to the robot position $\vec{x}_{t | t}$ as the next target waypoint $\vec{w}^*_{i+1} = \argmin_{\vec{w} \in \vec{w}_{[i+1, i+M]|i}} \lVert \vec{w} - \vec{x}_{t | t}\rVert_2$.
At each sampling iteration $j=0,\dots,M-1$, a new candidate $\vec{w}_{i+j+1|i}$ is drawn from the current error distribution $e_{i+j|i}(\vec{x})$. This distribution is then updated to $e_{i+j+1|i}(\vec{x})$ to incorporate the sampled waypoint, from which the next candidate is subsequently drawn. Figure~\ref{fig:ergodic_sampling} illustrates this sampling process.

%% file: sections/evaluation.tex
\section{Experimental Results}

\def\sceneRoom{\texttt{Single Office}}
\def\sceneRoomInline{\sceneRoom}

\def\sceneHallwayPartA{\texttt{Multi-Office}}
\def\sceneHallwayPartB{\texttt{+ Hallway}}
\def\sceneHallway{\sceneHallwayPartA\\\sceneHallwayPartB}
\def\sceneHallwayInline{\sceneHallwayPartA\ \sceneHallwayPartB}

\def\sceneKitchen{\texttt{Kitchen}}
\def\sceneKitchenInline{\sceneKitchen}

We evaluate our approach \textit{(i)}~in simulated closed-loop tests on the InteriorAgent~\cite{InteriorAgent2025} dataset, \textit{(ii)}~the Khronos~\cite{schmid_khronos_2024} mapping dataset, \textit{(iii)}~and on Hello Robot's Stretch 3 across two real-world office environments and further validated its generalization in a real-world, object-rich kitchen setting. The robot is equipped with an Orbbec Femto Bolt RGB-D camera and wirelessly connected to a workstation with an NVIDIA RTX 4090.
We first present results on datasets comparing directly to state-of-the-art prior work, followed by the real-world experiments.
A video showing our experimental results is available at \videolink{}.

\subsection{Evaluation on Public Datasets}

\subsubsection{Object-Goal Navigation Performance}

To evaluate our method against DynaMem~\cite{liu2024dynamem}, we design a set of $60$ object-goal navigation tasks in simulation, using eight synthetic interior scenes from the InteriorAgent dataset~\cite{InteriorAgent2025}. These tasks target semi-static environments, for which no standard closed-loop benchmark exists. Each task is categorized based on how the object-goal differs from a prior map generated via frontier exploration: \textit{(i)} the object-goal remains in the same location~(Known), \textit{(ii)} the object-goal was absent in the prior map~(Novel), and \textit{(iii)} the object-goal has moved~(Moved). To introduce changes such as object introduction and movement, we simply hide and reveal object instances of a particular class during the prior map generation phase and the task execution phase. For each task, methods are allocated \SI{15}{min} to generate a prior map; the scene is then updated, and methods have \SI{5}{min} to locate the target object. A trial is successful only if the robot reports that it has found the object and is within \SI{1.5}{m} of it. Table~\ref{tab:dynamem_compare} reports success rate and \ac{spl} \cite{evalspl2018}. Our method consistently outperforms DynaMem~\cite{liu2024dynamem} and the random-navigation baselines with respect to both metrics.

\begin{table}
    \centering
    \footnotesize
    \vspace{3mm}
    \caption{Object-Goal Navigation on the InteriorAgent~\cite{InteriorAgent2025} Dataset}\label{tab:dynamem_compare}\vspace{-3mm}%
    \centering
    \begin{tabular}{c@{\hspace{5pt}}|c@{\hspace{2pt}}c@{\hspace{7pt}}c@{\hspace{2pt}}c@{\hspace{7pt}}c@{\hspace{2pt}}c@{\hspace{5pt}}}
             & \multicolumn{2}{c}{Known} & \multicolumn{2}{c}{Novel} & \multicolumn{2}{c}{Moved} \\
     & SR $\uparrow$ & SPL $\uparrow$ & SR $\uparrow$ & SPL $\uparrow$ & SR $\uparrow$ & SPL $\uparrow$ \\
    \hline
    DynaMem \cite{liu2024dynamem} & $0.45$ & $0.36$ & $0.20$ & $0.11$ & $0.25$ & $0.24$ \\
    Random & $0.20$ & $0.10$ & $0.30$ & $0.14$ & $0.10$ & $0.02$ \\
    Ours & $\mathbf{0.65}$ & $\mathbf{0.62}$ & $\mathbf{0.45}$ & $\mathbf{0.35}$ & $\mathbf{0.50}$ & $\mathbf{0.43}$
    \end{tabular}\vspace{-2mm}
\end{table}

\subsubsection{Mapping in Changing Scenes}

We evaluate our mapping and change-detection performance on the Khronos dataset \cite{schmid_khronos_2024}. The dataset comprises two synthetic indoor environments--an apartment and a large multi-room office scene--providing RGB-D data, camera poses, ground-truth object annotations, object additions and removals, and human dynamics. Following Khronos, we report the same 4D extensions of precision, recall, and F1-score.

Note that we upper-bound the inconsistent F1 scores reported by Khronos~\cite{schmid_khronos_2024} with $\mathrm{F1} \leq \frac{1}{2}(\mathrm{Pre} + \mathrm{Rec})$, 
which also holds for the F1 extension to 4D. Our method detects changes with an on average $26.35\%$ higher F1 score, primarily because it incorporates semantic checks, whereas Khronos~\cite{schmid_khronos_2024} relies only on geometry. This difference becomes especially clear in the more cluttered apartment scene, where some objects provide only minimal geometric evidence for change. For instance, objects, such as a vase, are too small in scale for reliable geometric change detection, making semantic cues in the image essential. Although the system is not designed for handling dynamic objects, it still improves upon Khronos' Dynamics F1 score by $5.9\%$ on average.

\begin{table}
    \centering
    \footnotesize
    \caption{Results on the Khronos Dataset. Khronos' Results from \cite{schmid_khronos_2024}}\label{tab:khronos}
    \vspace{-4mm}%
$^+$: upper bounded by $\mathrm{F1} \leq \frac{1}{2}(\mathrm{Pre} + \mathrm{Rec})$ %

\begin{tabular}{c@{\hspace{5pt}}c@{\hspace{5pt}}|c@{\hspace{2pt}}c@{\hspace{2pt}}c@{\hspace{7pt}}|c@{\hspace{2pt}}c@{\hspace{2pt}}c@{\hspace{7pt}}}
 & Method & \multicolumn{3}{c}{Dynamics} & \multicolumn{3}{c}{Changes} \\

  &  & Pre & Rec & F1 & Pre & Rec & F1 \\

\hline
\multirow{2}{*}{\rotatebox{90}{\scriptsize apartment}} & Khronos~\cite{schmid_khronos_2024} & $90.4$ & $78.6$ & $84.1$ & $31.3$ & $69.1$ & $\leq50.2^+$ \\
 & \coloredited Ours & $\mathbf{92.1}$ & $\mathbf{86.1}$ & $\mathbf{88.9}$ & $\mathbf{94.8}$ & $\mathbf{84.8}$ & $\mathbf{89.3}$ \\
\\[0.5em]

\multirow{2}{*}{\rotatebox{90}{\scriptsize office}} & Khronos~\cite{schmid_khronos_2024} & $\mathbf{96.0}$ & $59.7$ & $73.2$ & $24.5$ & $\mathbf{54.2}$ & $\leq39.4^+$ \\
 & \coloredited Ours & $93.8$ & $\mathbf{71.3}$ & $\mathbf{80.2}$ & $\mathbf{66.5}$ & $47.0$ & $\mathbf{53.0}$ \\
\end{tabular}
\end{table}

\def\crop{0.4}
\begin{figure}
    \centering
    \begin{tikzpicture}
        \node (img1) {\includegraphics[width=0.4\linewidth, clip, trim={\crop cm, \crop cm, \crop cm, 0 cm}]{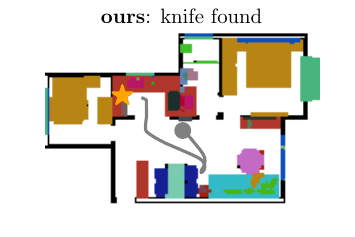}};
        \node[right=0.0cm of img1.south east, anchor=south west] (img2) 
            {\includegraphics[width=0.4\linewidth, clip, trim={\crop cm, \crop cm, \crop cm, 0 cm}]{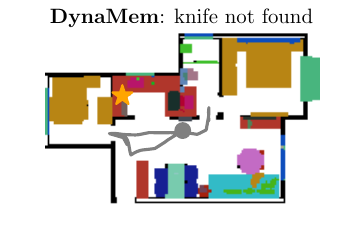}};
        \node[rotate=90] at ([xshift=-0.2cm]$(img1.north west)!0.5!(img1.south west)$) {\small Novel};
    \end{tikzpicture}\vspace{-3mm}

    \begin{tikzpicture}
        \node (img1) {\includegraphics[width=0.4\linewidth, clip, trim={\crop cm, \crop cm, \crop cm, 0 cm}]{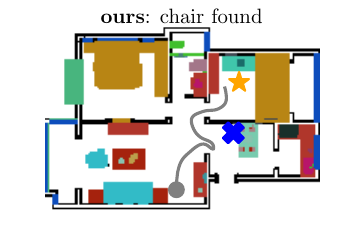}};
        \node[right=0.0cm of img1.south east, anchor=south west] (img2) 
            {\includegraphics[width=0.4\linewidth, clip, trim={\crop cm, \crop cm, \crop cm, 0 cm}]{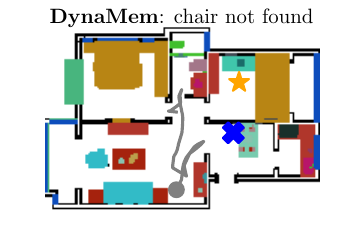}};
        \node[rotate=90] at ([xshift=-0.2cm]$(img1.north west)!0.5!(img1.south west)$) {\small Moved};
    \end{tikzpicture}\vspace{-5mm}
    \caption{Example of our method (left) and DynaMem \cite{liu2024dynamem} (right) searching for an unseen knife (top) and a moved chair (bottom). Our method checks the dining table, then kitchen and bedroom, while DynaMem explores randomly. Robot path and start shown in gray. Goal object marked with a yellow star; prior location with a blue cross.
    }\label{fig:example_dynamem}
\end{figure}

\subsection{Real World Evaluation}

\subsubsection{Qualitative Results}

\begin{figure}
    \centering
    \includegraphics[clip, trim={0cm, 0.5cm, 0cm, 4.0mm}]{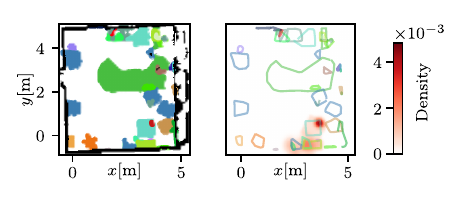}
    \vspace{-0.4cm}
    \caption[Exploration priority map (right) for finding a plate. A plate in this scene (left) is predicted to be found near a cup, coffee table, or cabinet. Some other objects are chairs and desks.]{A plate in an office scene (semantic map on the left) is predicted to be found near a
    cup \tikz{\draw[fill={rgb,255: red,198; green,23; blue,23}, draw=none]  circle(0.75ex);}, 
    coffee table \tikz{\draw[fill={rgb,255: red,103; green,216; blue,198}, draw=none]  circle(0.75ex);},
    or cabinet~\tikz{\draw[fill={rgb,255: red,166; green,228; blue,42}, draw=none]  circle(0.75ex);} according to the exploration priority map (right). Other objects in the scene include
    chairs \tikz{\draw[fill={rgb,255: red,61; green,126; blue,179}, draw=none]  circle(0.75ex);} and
    desks \tikz{\draw[fill={rgb,255: red,73; green,190; blue,65}, draw=none]  circle(0.75ex);}.
    }
    \label{fig:search_plate}
\end{figure}

To assess the semantic capabilities of the exploration priority map, we examine two scenarios: the maintenance task (Fig.~\ref{fig:heatmaps_maintain}) and the object-goal navigation task involving a plate (Fig.~\ref{fig:search_plate}). Figure~\ref{fig:heatmaps_maintain} shows a scene with several mapped objects likely to be moved (chairs and a ball) located outside the robot's current view. The synthetically injected changes cause the objects' stationarity scores to decay over time. As stationarity decreases, these areas gain relevance for the maintenance task, guiding the robot to revisit them. Thus, the robot continually identifies and revisits outdated regions. In Fig.~\ref{fig:search_plate}, the robot is located in a scene containing everyday objects (e.g., chairs, a cup, a coffee table, and a desk) and tasked to find a plate. The exploration priority map correctly infers the plate's likely location near the cup and coffee table. The supplementary video further shows the initial mapping, map maintenance, and object-goal navigation in a real-world \sceneKitchenInline{}, %
underscoring the applicability of our method to more object-rich environments.

\def\trimall{30mm, 4.2mm, 18mm, 3.5mm}
\begin{figure}
    \begin{tikzpicture}
        \node (scores) {\includegraphics[clip, trim={2mm, 0.5mm, 1.5mm, 11.5mm}]{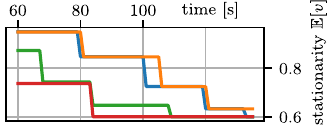}};
        \coordinate [above left=0.1cm and 5.15cm of scores.south east] (time0);
        \coordinate [above left=0.1cm and 3.8cm of scores.south east] (time1);
        \coordinate [above left=0.1cm and 2.6cm of scores.south east] (time2);
        \coordinate [above left=0.1cm and 1.9cm of scores.south east] (time3);
        
        \node[below left=-0.0cm and 0.0cm of scores.west, anchor=east] (scene) {\includegraphics[clip, trim={1mm, 3.8mm, 37.6mm, 1mm}]{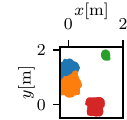}};

        \node[below=-0.25cm of time0] (init) {\includegraphics[clip, Trim={\trimall}]{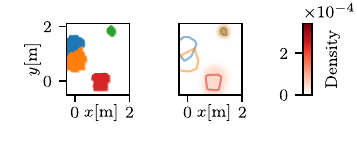}};
        \node[left=-0.31cm of init] (init_label)  {\includegraphics[clip, trim={1mm, 4.2mm, 48.7mm, 3.5mm}]{images/heatmap_examples/maintain_heatmap_initial.pdf}};

        \node[below=-0.25cm of time1] (2) {\includegraphics[clip, Trim={\trimall}]{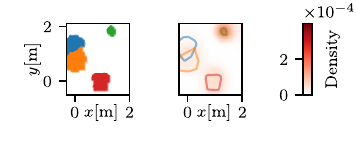}};
        \node[below=-0.25cm of time3] (4) {\includegraphics[clip, Trim={\trimall}]{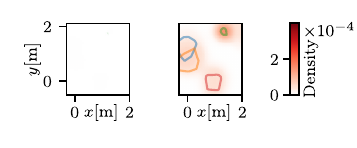}};
        \node[right=-0.0cm of 4] (colorbar) {\includegraphics[clip, Trim={45mm, 4.2mm, 0.5mm, 0.5mm}]{images/heatmap_examples/maintain_heatmap_final.pdf}};

        \coordinate [below=0.57cm of scores.north] (upper);

        \draw [dashed, thick] (time0 |- upper) -- (time0) -- ($(init.north) + (0, -0.15cm)$);
        \draw [dashed, thick] (time1 |- upper) -- (time1) -- ($(2.north) + (0, -0.15cm)$);
        \draw [dashed, thick] (time3 |- upper) -- (time3) -- ($(4.north) + (0, -0.15cm)$);
    \end{tikzpicture}
    \vspace{-1cm}
    \caption{Exploration priority map evolution over time while maintaining the map. The top left figure shows the map containing four objects. As the objects' stationary score decays (top right), they become more relevant to the map-maintenance task, which is reflected in the task likelihood maps~(bottom, from left to right).}
    \label{fig:heatmaps_maintain}
    \vspace{-0.8em}
\end{figure}

\def\trimall{0.3cm, 1cm, 0.2cm, 0.75cm}
\def\trimlast{0.3cm, 0.2cm, 0.2cm, 0.75cm}

\input{sections/geometric_comparison_table.tex}

\subsubsection{Semi-Static Mapping Accuracy}

We evaluate mapping accuracy in two environments: a single office and a two-room layout connected by a hallway. Ground truth is obtained via high-precision LiDAR scans and manual object labeling, both before and after scene changes were introduced by moving, adding, and removing objects.

For geometric accuracy, ground truth and mapped point clouds are voxelized at a resolution of \SI{0.1}{m}, overlaid, and analyzed for true/false positives and negatives. To isolate the impact of changed objects, we limit the comparison to the bounding boxes of moved objects (see Fig.~\ref{fig:geometric_comparison}). We compare against three baselines: frontier exploration of the initial scene, full remapping of the changed scene, and a no-change-detection method that cannot update mapped objects. As shown in Tab.~\ref{tab:geom_accuracy}, the precision drops slightly from initial mapping to updated maps, with a larger drop in the \sceneHallwayInline{} scenario. Our method performs on par with full remapping and consistently outperforms the no-change-detection method, which exhibits low precision and high false positive rates from failing to remove moved or deleted objects, as depicted in Fig.~\ref{fig:geometric_comparison}.

Object detection accuracy is measured by comparing mapped object instances using precision, recall, and F1 scores~(Tab.~\ref{tab:semantic_accuracy}). Note that we omit full map rebuilding from this comparison. These results align with the geometric results: no-change-detection yields low precision, whereas our method strikes a balance between precision and recall, as shown by the F1 score. Notably, our method slightly improves over the initial mapping due to more observation data being available.

\def\trimall{9.0cm, 11cm, 13cm, 4cm}
\begin{figure}
    \centering
    \makebox[0.5\textwidth]{\begin{tikzpicture}
        \node [anchor=south west] (img2) {\includegraphics[width=1.6cm, clip, Trim={\trimall}]{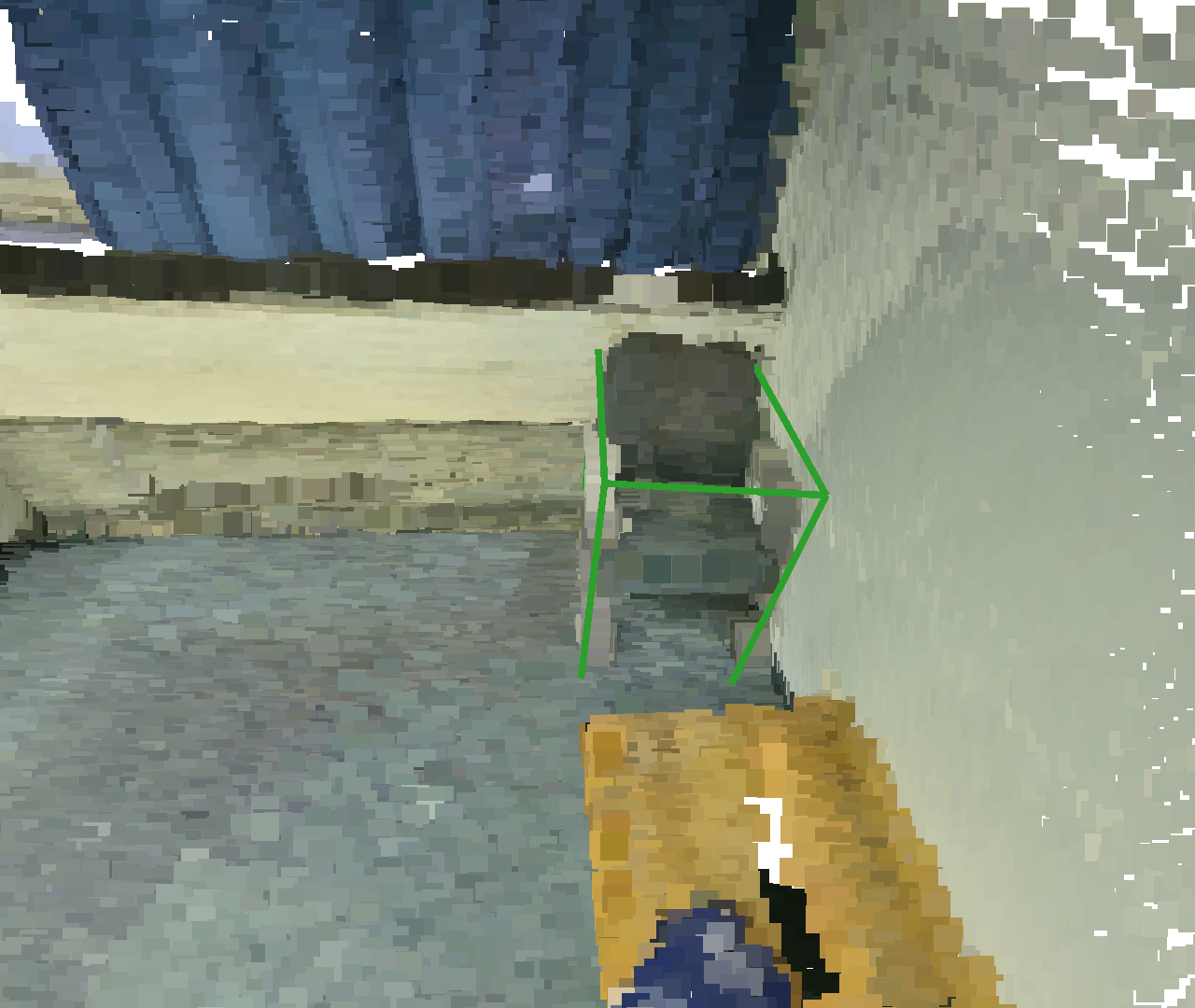}};
        
        \node [anchor=south west, left=0.2cm of img2] (img1) {\includegraphics[width=1.6cm, clip, Trim={\trimall}]{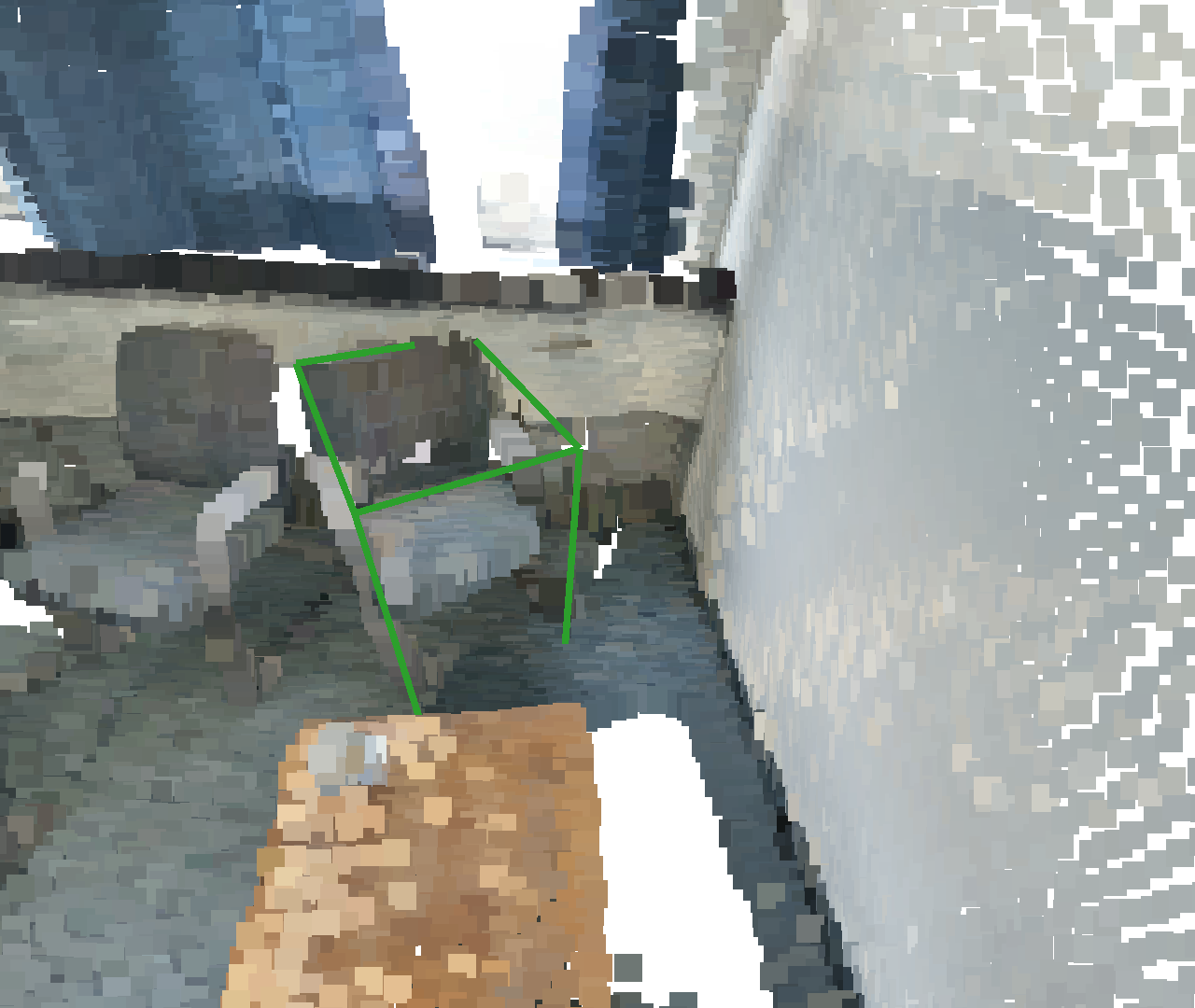}};

        \def\trimall{9.0cm, 11cm, 13cm, 11.5cm}
        \node [anchor=south west, above right=0.5cm and -0.2cm of img2.south east] (img2_1) {\includegraphics[width=1.6cm, clip, Trim={\trimall}]{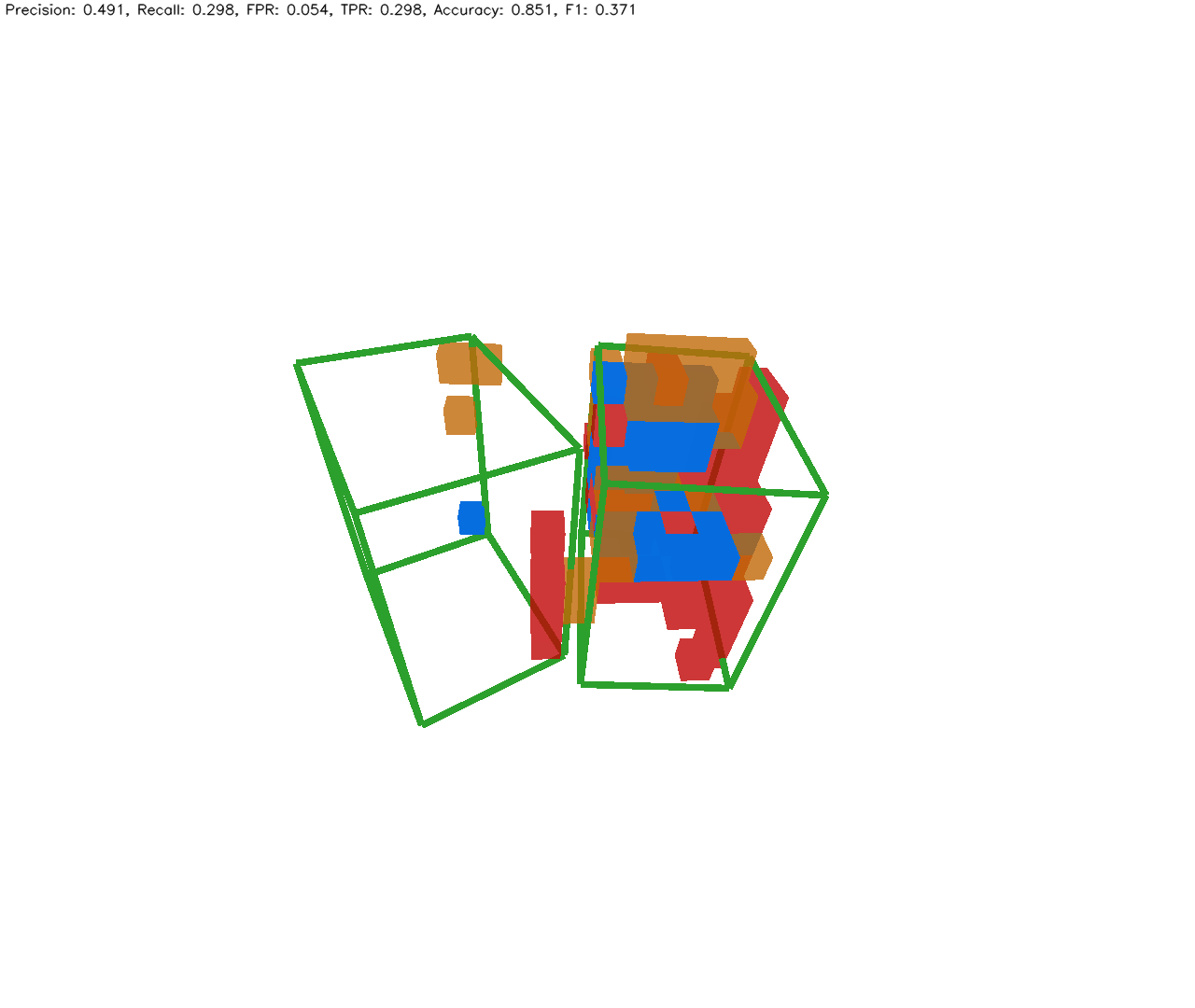}};
        \node [anchor=south west, right=-0.2cm of img2_1.east] (img2_2) {\includegraphics[width=1.6cm, clip, Trim={\trimall}]{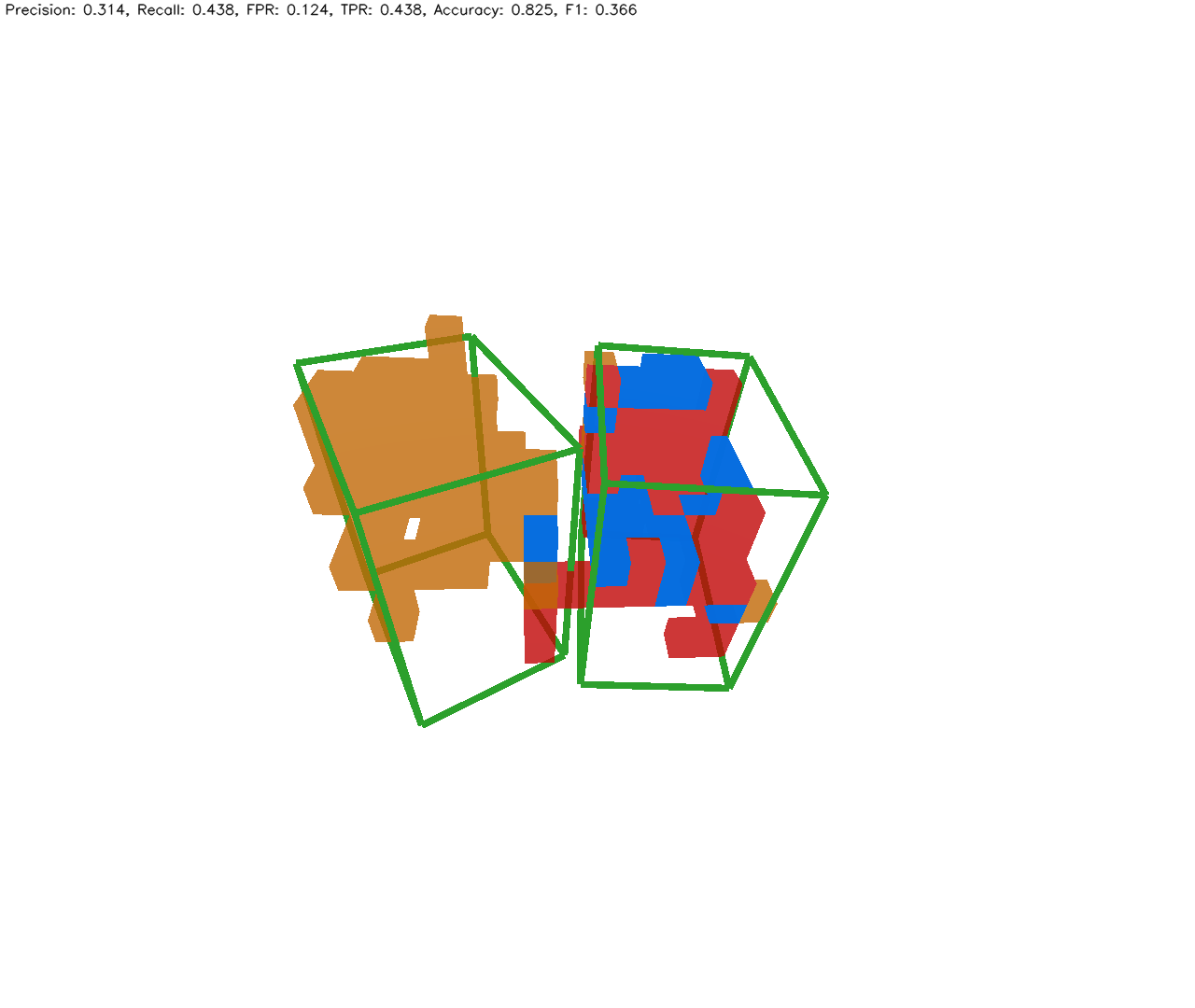}};
        \node [anchor=south west, right=-0.2cm of img2_2.east] (img2_3) {\includegraphics[width=1.6cm, clip, Trim={\trimall}]{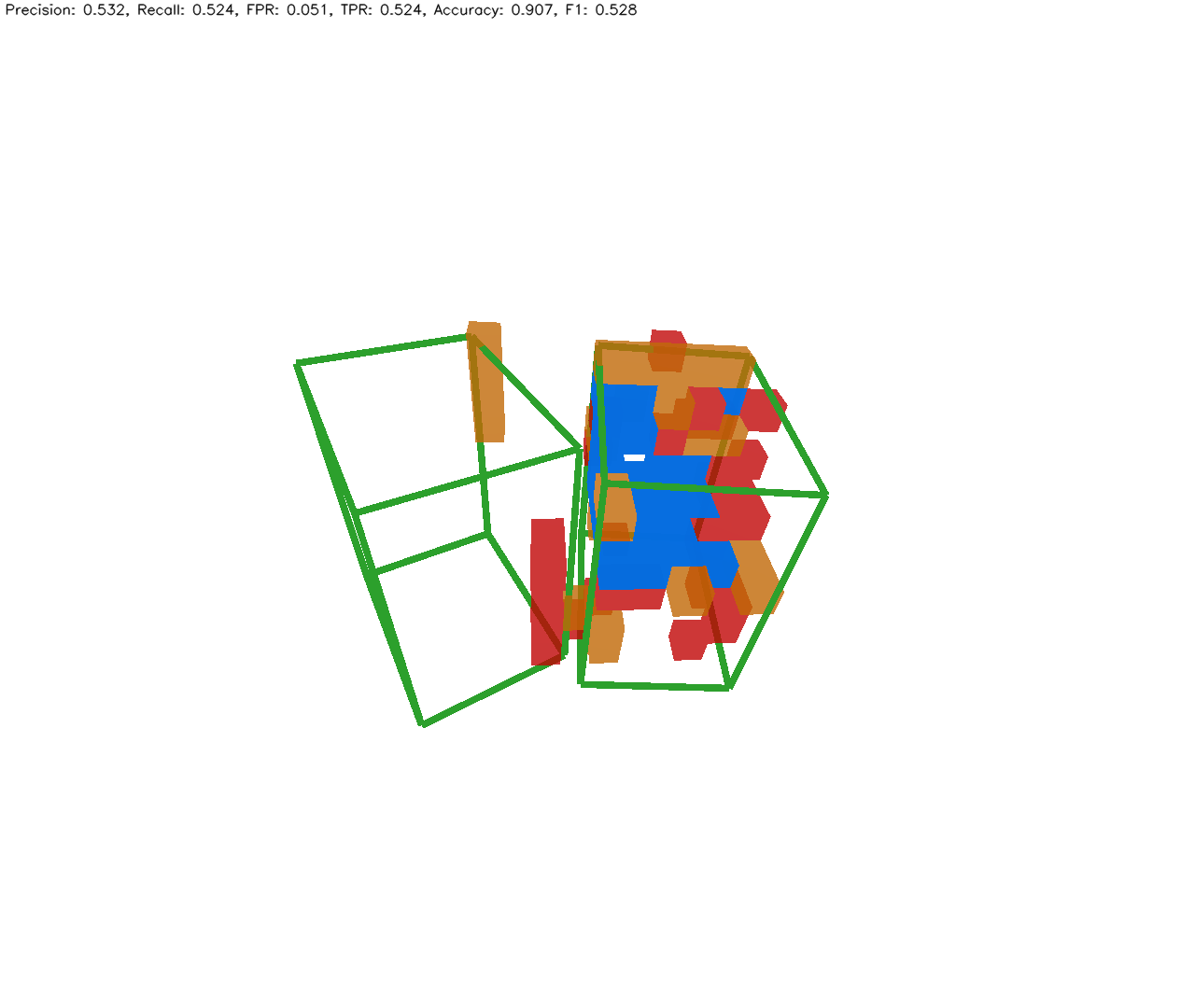}};

        \node [below=0.05cm of img1.south, rotate=0, font=\footnotesize, anchor=center, align=center] {Initial State};
        \node [below=0.05cm of img2.south, rotate=0, font=\footnotesize, anchor=center, align=center] {Updated State};

        \node [below=-0.1cm of img2_1.south, font=\footnotesize, anchor=north, align=center] {Full\\Rebuild};
        \node [below=-0.1cm of img2_2.south, font=\footnotesize, anchor=north, align=center] {wo/ Change\\Detection};
        \node [below=-0.1cm of img2_3.south, font=\footnotesize\bfseries, anchor=north, align=center] {w/ Change\\Detection};

        \draw ($(img1.north east)!0.5!(img2.north west) + (0,-2mm)$) 
      -- ($(img1.south east)!0.5!(img2.south west) + (0,2mm)$);

      \node [right=-1mm of img2_3.east, rectangle, fill=white] {};
    \end{tikzpicture}}
    \vspace{-0.7cm}
    \caption[]{Voxel-wise comparison to the groundtruth of a mapped chair at initial (left) and updated (right) scene state. The voxels show
    true positives \tikz{\draw[fill={rgb,255: red,6; green,108; blue,221}, draw=none]  (0,0) rectangle +(1.5ex,1.5ex);},
    false positives \tikz{\draw[fill={rgb,255: red,205; green,135; blue,56}, draw=none]  (0,0) rectangle +(1.5ex,1.5ex);},
    and false negatives \tikz{\draw[fill={rgb,255: red,205; green,56; blue,56}, draw=none]  (0,0) rectangle +(1.5ex,1.5ex);}.
    The method without change detection (right second) fails to remove and update the chair, resulting in many false positives.
    }\label{fig:geometric_comparison}
\end{figure}

\subsubsection{Semantic Exploration Efficiency}

We evaluate our exploration approach against two baselines: a random policy that selects reachable waypoints at random, and a patrol policy that follows a~$\SI{2}{\metre}$ grid pattern.

For map maintenance, we measure the ratio of detected to applied changes within a fixed time, counting additions and removals separately (moved objects count as both). Table~\ref{tab:addition_removal} shows that in the single-room \sceneRoomInline{} scenario, all methods perform similarly, but in the larger \sceneHallwayInline{} scenario, our method outperforms both by quickly passing through the hallway to reach areas with more changes.

Finally, we evaluate our method's effectiveness in locating previously unseen objects. Starting from an initially mapped scene, we place a new object for each run and task the robot with finding it within a maximum allowed time of $\SI{2}{\min}$ $\SI{30}{\second}$. Success requires the robot to stop near and face the target object. Each method is tested in eight runs: $2\times$ books on a shelf, $2\times$ books on a chair, $2\times$ a bowl on a coffee table, and $2\times$ a keyboard on a desk. Table~\ref{tab:searching} reports the success rate~$r_{\text{s}}$, mean success time~$t_{\text{s}}$, and weighted mean success time~$\frac{t_{\text{s}}}{r_{\text{s}}}$. While the random strategy can be faster, it has a success rate of only $25\%$. Both our method and the patrol strategy are consistently reliable, with ours being $\sim$14\% faster while maintaining a similar success rate for object goal navigation.

\subsection{Ablation and Sensitivity Analysis}

\begin{figure}
    \vspace{-4mm} %
    \centering%
    \makebox[0.4\textwidth]{%
    \subfloat[The stationarity rises and drops when objects appear and disappear.]{%
    \footnotesize%
    \centering%
    \def\trimall{0mm, 4.5cm, 0, 4cm}%
    \begin{tikzpicture}
        \node (graph) at (0,0) {\includegraphics[clip, trim={2mm, 0mm, 0, 5mm}]{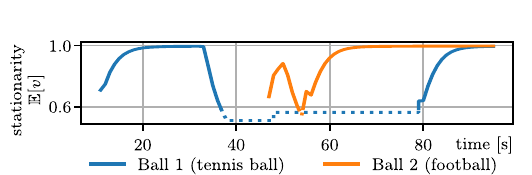}};
        
        \coordinate [above=-1mm of graph] (ypos);
        \coordinate (x0) at (-2.85,-0.37);
        \coordinate (x1) at (-1.15,-0.37);
        \coordinate (x2) at (-0.05,-0.37);
        \coordinate (x3) at (2.45,-0.37);
        \node[anchor=south, yshift=-1mm] at (x0 |- ypos) {\includegraphics[width=1cm,clip,Trim={\trimall}]{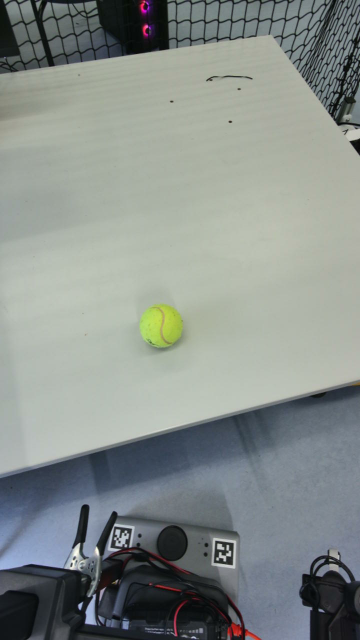}};
        \node[anchor=south, yshift=-1mm] at (x1 |- ypos) {\includegraphics[width=1cm,clip,Trim={\trimall}]{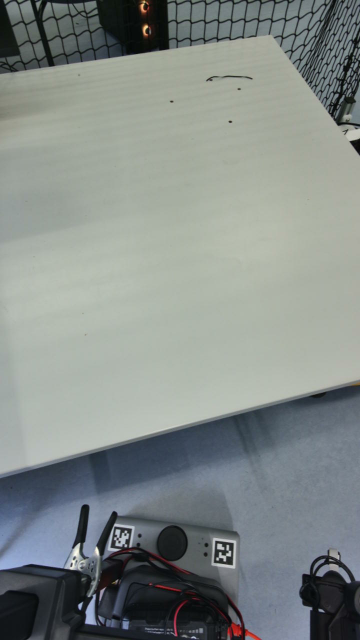}};
        \node[anchor=south, yshift=-1mm] at (x2 |- ypos) {\includegraphics[width=1cm,clip,Trim={\trimall}]{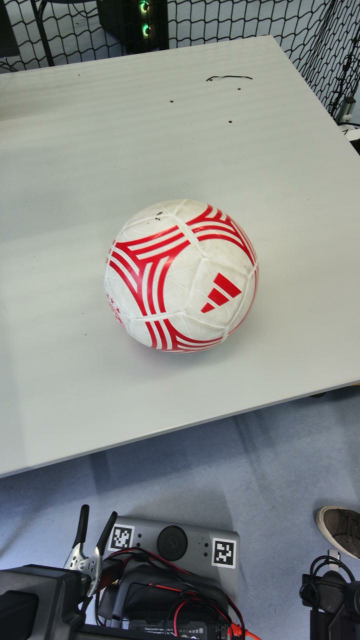}};
        \node[anchor=south, yshift=-1mm] at (x3 |- ypos) {\includegraphics[width=1cm,clip,Trim={\trimall}]{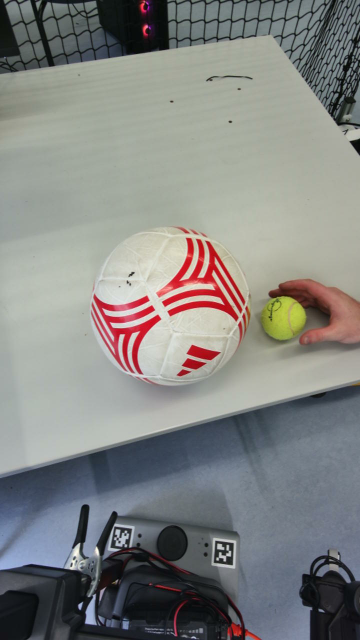}};

        \draw[dashed, thick, {Latex[width=1.5mm,length=1.5mm]}-] (x0 |- ypos) -- (x0);
        \draw[dashed, thick, {Latex[width=1.5mm,length=1.5mm]}-] (x1 |- ypos) -- (x1);
        \draw[dashed, thick, {Latex[width=1.5mm,length=1.5mm]}-] (x2 |- ypos) -- (x2);
        \draw[dashed, thick, {Latex[width=1.5mm,length=1.5mm]}-] (x3 |- ypos) -- (x3);

        \coordinate [above right=1.14cm and 0.22cm of x2] (start);
        \coordinate (end) at ($ (start) + (-72:0.7cm) $);
        \draw[-{Latex[width=1.5mm,length=1.5mm]}, thick, color={rgb,255:red,255; green,127; blue,14}] ($ (start) + (1mm,1mm) $) -- ($ (end) + (1mm,1mm) $) node [fill=white, inner sep=0pt, pos=0.2, anchor=east, xshift=-1mm, yshift=0.3mm, font=\scriptsize, align=right] {Ball rolling};
    \end{tikzpicture}\label{fig:object_translated_and_merged}%
    }}%
    
    \centering
    \subfloat[The table shows which similarity measures are necessary to successfully reidentify the shown object instances. Only when both measures are used all three cases are covered.]{%
    \def\width{0.06}
    \footnotesize
    \begin{tabular}{c|c c c}
    \diagbox[height=1.9cm]{Enabled\\\makebox[0pt][l]{(i.e., $\tau_{\cdot}>0$)}\\\makebox[0pt][l]{Similarity Measures}}{Objects to\\Reidentify} & \makecell{Different\\\makebox[40pt]{Sem. \& Geom.}\\\includegraphics[width=\width\textwidth]{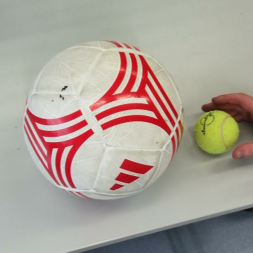}} & \makecell{Different\\Semantics\\\includegraphics[width=\width\textwidth]{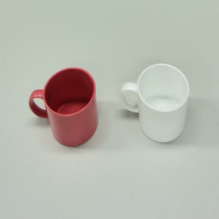}} & \makecell{Different\\Geometries\\\includegraphics[width=\width\textwidth]{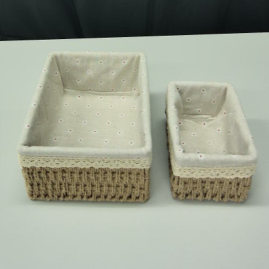}} \\
    \hline
    $\tau_{\mathrm{geo}} > 0, \tau_{\mathrm{sem}} = 0$ & \ding{51} & \ding{55} & \ding{51} \\
    $\tau_{\mathrm{geo}} = 0, \tau_{\mathrm{sem}} > 0$ & \ding{51}  & \ding{51} &\ding{55} \\
    \coloredited $\mathbf{\tau_{\mathrm{geo}} > 0, \tau_{\mathrm{sem}} > 0}$ & \coloredited \ding{51} & \coloredited \ding{51} & \coloredited \ding{51}
    \end{tabular}
    }
  \caption{Example of removal, reintroduction, and translation. Our system can distinguish between objects of the same class and reidentify previously shown object instances.}
  \label{fig:reidentificaion}
\end{figure}

We conduct an ablation study to evaluate the contribution of the semantic and geometric similarity measures (\eqref{eq:similarity_sem} and \eqref{eq:similarity_geo}) to our approach's ability to track object instances over time. We place, remove, and reintroduce two object instances of the same class in front of the robot as shown in Fig.~\ref{fig:object_translated_and_merged}, and verify whether our system can correctly distinguish the two instances from and each other while still correctly identifying the removed and later reintroduced object. For the two balls, which are both geometrically (their radius) and semantically (their color) different, either similarity measure alone suffices for successful reidentification. For other object pairs (such as the cups and baskets), which differ only in either semantics or geometry, the respective similarity measure must be enabled to distinguish them. Conclusively, the integration of both similarity measures is necessary to robustly track object instances in a semi-static setting.

We analyze our method's sensitivity to parameters on the Khronos~\cite{schmid_khronos_2024} dataset. Varying $\tau_{\mathrm{sem}}$, $\tau_{\mathrm{geo}}$, $\theta_{\mathrm{r}}$, and $d_{\mathrm{ICP}}$ over $[0.1,0.99]$, $[0.1,0.99]$, $[0.1,0.5]$, and $[0.0001,0.1]$, respectively, results in a maximum F1 score change of $\pm 7.0$. Overall, the method is fairly robust to parameter changes. We recommend starting with lenient $\tau_{\mathrm{sem}}$, $\tau_{\mathrm{geo}}$, and $d_{\mathrm{ICP}}$ values and increasing them cautiously to avoid false positive detections.

\subsection{Computation Performance}

\begin{figure}
  \centering
  \includegraphics[clip, trim={2mm, 2mm, 0, 1mm}]{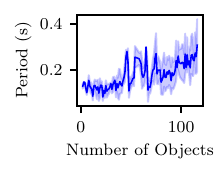}
  \hfill
  \input{figures/performance_chart.tex}
  \vspace{-2mm}
  \caption{Map update interval against the number of objects $\lvert\gls{obj_lib}\rvert$ (left), and relative processing time spent on main mapping components (right).%
  }\label{fig:timing}
\end{figure}

Figure~\ref{fig:timing} (left) indicates that map-update time grows with object count due to per-object operations and the progressively growing background point cloud. However, even with 100 objects, the map updates take only $\SI{0.2}{\second}$ on average, showing the real-time capability of our approach. A detailed decomposition of the computation is further included in the right panel of Fig.~\ref{fig:timing}. %

\vspace{-1mm}

%% file: sections/geometric_comparison_table.tex
\def\officeBkgInitP{} \def\officeBkgInitR{} \def\officeBkgInitAccuracy{} \def\officeBkgInitFPR{}
\def\officeBkgMaintainP{} \def\officeBkgMaintainR{} \def\officeBkgMaintainAccuracy{} \def\officeBkgMaintainFPR{}
\def\officeBkgNoP{} \def\officeBkgNoR{} \def\officeBkgNoAccuracy{} \def\officeBkgNoFPR{}
\def\hallwayBkgInitP{} \def\hallwayBkgInitR{} \def\hallwayBkgInitAccuracy{} \def\hallwayBkgInitFPR{}
\def\hallwayBkgMaintainP{} \def\hallwayBkgMaintainR{} \def\hallwayBkgMaintainAccuracy{} \def\hallwayBkgMaintainFPR{}
\def\hallwayBkgNoP{} \def\hallwayBkgNoR{} \def\hallwayBkgNoAccuracy{} \def\hallwayBkgNoFPR{}

\def\hallwayInitP{75.3} \def\hallwayInitR{34.8} \def\hallwayInitFPR{1.9} \def\hallwayInitTPR{34.8} \def\hallwayInitAccuracy{89.0} \def\hallwayInitF{47.6}
\def\hallwayMaintainP{38.0} \def\hallwayMaintainR{30.0} \def\hallwayMaintainFPR{5.6} \def\hallwayMaintainTPR{30.0} \def\hallwayMaintainAccuracy{87.8} \def\hallwayMaintainF{33.6}
\def\hallwayNoP{30.7} \def\hallwayNoR{44.3} \def\hallwayNoFPR{12.0} \def\hallwayNoTPR{44.3} \def\hallwayNoAccuracy{83.4} \def\hallwayNoF{36.3}
\def\hallwayResetP{40.0} \def\hallwayResetR{32.8} \def\hallwayResetFPR{6.2} \def\hallwayResetTPR{32.8} \def\hallwayResetAccuracy{86.9} \def\hallwayResetF{36.0}

\def\officeInitP{75.9} \def\officeInitR{32.4} \def\officeInitFPR{2.0} \def\officeInitTPR{32.4} \def\officeInitAccuracy{87.1} \def\officeInitF{45.4}
\def\officeMaintainP{71.2} \def\officeMaintainR{27.8} \def\officeMaintainFPR{2.1} \def\officeMaintainTPR{27.8} \def\officeMaintainAccuracy{87.1} \def\officeMaintainF{39.9}
\def\officeNoP{41.0} \def\officeNoR{43.7} \def\officeNoFPR{10.5} \def\officeNoTPR{43.7} \def\officeNoAccuracy{83.0} \def\officeNoF{42.3}
\def\officeNoP{44.1} \def\officeNoR{40.8} \def\officeNoFPR{9.2} \def\officeNoTPR{40.8} \def\officeNoAccuracy{83.2} \def\officeNoF{42.5}
\def\officeResetP{74.1} \def\officeResetR{24.9} \def\officeResetFPR{1.8} \def\officeResetTPR{24.9} \def\officeResetAccuracy{85.8} \def\officeResetF{37.2}

\begin{table}
    \centering
    \footnotesize
    \vspace{2.5mm} %
    \caption[Geometric Accuracy Pre and Post Scene Changes]{Geometric Accuracy Pre and Post Scene Changes}\label{tab:geom_accuracy}%
    \vspace{-3mm}%
    \begin{tabular}{c|c@{\hspace{5pt}}cc@{\hspace{5pt}}cc@{\hspace{5pt}}c}
         Use Change                      & \multicolumn{2}{c}{Precision $\uparrow$} & \multicolumn{2}{c}{Accuracy $\uparrow$} & \multicolumn{2}{c}{FPR $\downarrow$} \\
         Detection                       & A & B       &  A  &    B          &   A   &   B       \\
        \hline
        
         \coloredited Initial Exploration & $\officeInitP$  & $\hallwayInitP$    & $\officeInitAccuracy$  & $\hallwayInitAccuracy$   & $\officeInitFPR$ & $\hallwayInitFPR$    \\[0.33em]
         Full Rebuild                    & $\mathbf{\officeResetP}$    & $\mathbf{\hallwayResetP}$      & $\officeResetAccuracy$             & $\hallwayResetAccuracy$              & $\mathbf{\officeResetFPR}$   & $\hallwayResetFPR$               \\
         \xmark                          & $\officeNoP$                & $\hallwayNoP$                  & $\officeNoAccuracy$                & $\hallwayNoAccuracy$                 & $\officeNoFPR$               & $\hallwayNoFPR$                  \\
         \cmark \hspace{.25ex} (Ours)    & $\officeMaintainP$          & $\hallwayMaintainP$            & $\mathbf{\officeMaintainAccuracy}$ & $\mathbf{\hallwayMaintainAccuracy}$  & $\officeMaintainFPR$         & $\mathbf{\hallwayMaintainFPR}$
    \end{tabular}

    \vspace{0.5em}A: \sceneRoomInline{}, \quad B: \sceneHallwayInline{}
\end{table}

%% file: figures/performance_chart.tex
\begin{tikzpicture}[font=\scriptsize]
    \definecolor{mpl_blue}{RGB}{31,119,180}
    \definecolor{mpl_orange}{RGB}{255,127,14}
    \definecolor{mpl_green}{RGB}{44,160,44}
    \definecolor{mpl_red}{RGB}{214,39,40}
    \definecolor{mpl_purple}{RGB}{148,103,189}
    \definecolor{mpl_brown}{RGB}{140,86,75}
    \definecolor{mpl_pink}{RGB}{227,35,200}
    \definecolor{mpl_gray}{RGB}{127,127,127}
    \definecolor{mpl_olive}{RGB}{188,189,34}
    \definecolor{mpl_cyan}{RGB}{23,190,207}
    \pie[
        radius=1,           %
        sum=auto,           %
        after number = \%,    %
          color={
              mpl_blue!40!white,           %
              mpl_orange!40!white,          %
              mpl_green!40!white,         %
              mpl_red!40!white,        %
              mpl_gray!20!white%
          },
        hide number,
        draw=white
    ]{
        35.52/,%
        33.29/,%
        20.64/,%
        6.23/,%
        4.32/%
    }

    \node at (60:5mm) {$35.5\%$};
    \node at (185:5mm) {$33.3\%$};
    \node at (291:6mm) {$20.6\%$};

      \begin{scope}[xshift=1.2cm, yshift=0.7cm]
        \foreach \i/\col/\label/\val in {
          0/mpl_blue!40!white/\shortstack[l]{Detecting Object\\[-0.3em]Candidates $\gls{detections}_t$}/,
          1/mpl_orange!40!white/\shortstack[l]{Compute Expected\\[-0.3em]Objects $\gls{obj_lib}_{t,\mathrm{exp}}$}/,
          2/mpl_green!40!white/\shortstack[l]{Object Matching\\[-0.3em]and Merging}/,
          3/mpl_red!40!white/\shortstack[l]{Background\\[-0.3em]Pointcloud}/,
          4/mpl_gray!20!white/Other/
        }{
            \draw[fill=\col, draw=none] (0,-\i*0.55) rectangle (0.3,-\i*0.55+0.3);
          \node[right, xshift=-0.2cm] at (0.6,-\i*0.55+0.125) {\label};
        }
      \end{scope}
\end{tikzpicture}

%% file: sections/conclusion.tex
\section{Conclusion}

In this work, we proposed a novel open-vocabulary semantic exploration approach for robots operating in semi-static environments. Beyond traditional object-goal navigation, our approach actively targets map regions likely to be outdated. We verified its effectiveness against state-of-the art methods, achieving $25\%$ higher \ac{spl} and $26\%$ increased change detection F1 in changing scenes. Further, we confirm its transferability to real-world scenarios.

\input{sections/real_world_result_tables.tex}

%% file: sections/real_world_result_tables.tex
\begin{table*}
\begin{minipage}[t]{0.35\textwidth}
    \caption{Object Detection Accuracy Pre and Post Scene Changes}\label{tab:semantic_accuracy}
    \centering
    \vspace{-2mm}%
    A: \sceneRoomInline{} (25 objects), \quad B: \sceneHallwayInline{} (34 objects)\vspace{0.5em}
    \begin{tabular}{c@{\hspace{5pt}}|c@{\hspace{2pt}}c@{\hspace{7pt}}c@{\hspace{2pt}}c@{\hspace{7pt}}c@{\hspace{2pt}}c@{\hspace{5pt}}}
                                        Use Change            & \multicolumn{2}{c}{Precision $\uparrow$} & \multicolumn{2}{c}{Recall $\uparrow$} & \multicolumn{2}{c}{$\mathrm{F_1}\uparrow$} \\
                                        Detection             & A &        B                              & A & B                                & A & B \\
        \hline
                                        \coloredited\makecell{Initial Expl.} & 0.85 & 0.72      & 0.68 & 0.68  & 0.76 &  0.70    \\[0.33em]
                                        \xmark                          & 0.39             & 0.36                  & \bfseries 0.84   & \bfseries 0.68    & 0.53             &  0.47                \\
                                        \cmark \hspace{.25ex} (Ours)    & \bfseries 0.90   & \bfseries 0.82        & 0.72             & \bfseries 0.68    & \bfseries 0.80   &  \bfseries 0.74 
    \end{tabular}
\end{minipage}%
\hfill
\begin{minipage}[t]{0.32\textwidth}
    \caption{Change Detection Efficiency}\label{tab:addition_removal}
    \vspace{-2mm}%
    A: \sceneRoomInline{} ($\SI{5}{\min}$), \quad B: \sceneHallwayInline{} ($\SI{10}{\min}$)\vspace{0.87em}
    \centering
    \begin{tabular}{r@{\hspace{5pt}}|c@{\hspace{2pt}}c@{\hspace{7pt}}c@{\hspace{2pt}}c@{\hspace{5pt}}}
    & \multicolumn{2}{c}{Identified} & \multicolumn{2}{c}{Identified} \\
    & \multicolumn{2}{c}{Additions ($\%$)} & \multicolumn{2}{c}{Removals ($\%$)} \\
    & A & B & A & B \\
    \hline
    Patrol    & $90.9$            & $25.0$            & $81.8$           & $0.0$             \\
    Random    & $\mathbf{100.0}$  & $37.5$            & $90.9$           & $40.0$ \\ %
    Ours      & $90.9$            & $\mathbf{87.5}$   & $\mathbf{100.0}$ & $\mathbf{100.0}$ 
    \end{tabular}
\end{minipage}%
\hfill
\begin{minipage}[t]{0.32\textwidth}
    \caption{Object Goal Navigation Performance}\label{tab:searching}
    \centering\vspace{-0.5em}
    \vspace{-2mm}%
    Eight Runs Each ($2\times$ books on shelf, $2\times$ books on chair, $2\times$ bowl on coffee table, $2\times$ keyboard on desk)\vspace{0.2em}
    \centering
    \begin{tabular}{r@{\hspace{5pt}}|c@{\hspace{5pt}}c@{\hspace{5pt}}c@{\hspace{5pt}}}
           & \makecell{Success\\Rate\\($\%$)}$\uparrow$ & \makecell{Mean\\Success\\Time (s)}$\downarrow$ & \makecell{Weighted\\Success\\Time (s)}$\downarrow$ \\
        \hline                                                                                                                                                                  
        Patrol   & \bfseries 100                 & 73.75                    & 73.75            \\
        Random   & 25                  & \bfseries 41.50                    & 166.00           \\   
        Ours     & \bfseries 100                 &  62.88                    & \bfseries 62.88
    \end{tabular}
\end{minipage}
\vspace{-5mm}
\end{table*}